\documentclass[conference]{IEEEtran}
\usepackage{fancyhdr}
\usepackage{tikz}
\usepackage{amsmath, amsfonts, amssymb}
\usepackage{microtype}
\usepackage{graphicx}
\usepackage{subcaption}
\usepackage{listings}
\usepackage{adjustbox}
\usepackage{booktabs}
\usepackage{hyperref}
\usepackage{multirow}
\usepackage{xcolor}
\usepackage{algorithmic}
\usepackage[ruled,linesnumbered]{algorithm2e}
\usepackage{comment}

\definecolor{codegreen}{rgb}{0.0,0.5,0.0}
\definecolor{codegray}{rgb}{0.5,0.5,0.5}
\definecolor{codepurple}{rgb}{0.58,0,0.82}
\definecolor{codeblue}{rgb}{0.0, 0.5, 1.0}
\lstdefinestyle{mystyle}{
    basicstyle={\footnotesize\ttfamily},
    commentstyle=\color{codegreen},
    keywordstyle=\color{black},
    numberstyle=\tiny\color{codegray},
    breakatwhitespace=false,
    captionpos=b,
    keepspaces=true,
    numbers=left,
    numbersep=5pt,
    showspaces=false,
    showstringspaces=false,
    showtabs=false,
    tabsize=2
}
\newcommand{\showcomments}{yes}

\newcommand{\final}[1]{
   \unskip{\textcolor{black}{#1}}\xspace
}
\newcommand{\revision}[1]{
   \unskip{\textcolor{black}{#1}}\xspace
}
\newcommand{\fixme}[1]{
    \ifthenelse{\equal{\showcomments}{yes}}{\textcolor{red}{[#1]}}{\ignorespaces}
}
\newcommand\yida[1]{
    \ifthenelse{\equal{\showcomments}{yes}}{\textcolor{red}{[yida: #1]}}{\ignorespaces}
}
\newcommand\mli[1]{
    \ifthenelse{\equal{\showcomments}{yes}}{\textcolor{orange}{[mu: #1]}}{\ignorespaces}
}
\newcommand\zd[1]{
    \ifthenelse{\equal{\showcomments}{yes}}{\textcolor{brown}{[zd: #1]}}{\ignorespaces}
}
\newcommand\mj[1]{
    \ifthenelse{\equal{\showcomments}{yes}}{\textcolor{magenta}{[mj: #1]}}{\ignorespaces}
}
\newcommand\zz[1]{
    \ifthenelse{\equal{\showcomments}{yes}}{\textcolor{purple}{[zz: #1]}}{\ignorespaces}
}
\newcommand\reply[1]{
    \ifthenelse{\equal{\showcomments}{yes}}{\textcolor{magenta}{[reply: #1]}}{\ignorespaces}
}

\newcommand{\system}[0]{FeatGraph\xspace}

\DeclareMathAlphabet{\mathcal}{OMS}{cmsy}{m}{n}

\title{{\system}: A Flexible and Efficient Backend for Graph Neural Network Systems}

\IEEEoverridecommandlockouts

\author{
    \IEEEauthorblockN{Yuwei Hu\IEEEauthorrefmark{2}$^*$,
                      Zihao Ye\IEEEauthorrefmark{4},
                      Minjie Wang\IEEEauthorrefmark{4},
                      Jiali Yu\IEEEauthorrefmark{4},
                      Da Zheng\IEEEauthorrefmark{4},
                      Mu Li\IEEEauthorrefmark{4},
                      Zheng Zhang\IEEEauthorrefmark{4},
                      Zhiru Zhang\IEEEauthorrefmark{2},
                      Yida Wang\IEEEauthorrefmark{4}}
    \\
    \IEEEauthorrefmark{2}School of ECE, Cornell University; \{yh457, zhiruz\}@cornell.edu\\
    \IEEEauthorrefmark{4}Amazon Web Services; \{yeziha, minjiw, yjial, dzzhen, mli, zhaz, wangyida\}@amazon.com
    \thanks{$^*$Work done during an internship at Amazon Web Services.}
}

\begin{document}

\maketitle

\thispagestyle{fancy}
\lhead{}
\rhead{}
\chead{}
\lfoot{\footnotesize{
SC20, November 9-19, 2020, Is Everywhere We Are
\newline 978-1-7281-9998-6/20/\$31.00 \copyright 2020 IEEE}}
\rfoot{}
\cfoot{}
\renewcommand{\headrulewidth}{0pt}
\renewcommand{\footrulewidth}{0pt}

\begin{abstract}

%%% camera-ready requires maximum 150 words
Graph neural networks (GNNs) are gaining popularity as a promising approach to machine learning on graphs. Unlike traditional graph workloads where each vertex/edge is associated with a scalar, GNNs attach a feature tensor to each vertex/edge. This additional feature dimension, along with consequently more complex vertex- and edge-wise computations, has enormous implications on locality and parallelism, which existing graph processing systems fail to exploit.

This paper proposes {\system} to accelerate GNN workloads by co-optimizing graph traversal and feature dimension computation. {\system} provides a flexible programming interface to express diverse GNN models by composing coarse-grained sparse templates with fine-grained user-defined functions (UDFs) on each vertex/edge. \system~incorporates optimizations for graph traversal into the sparse templates and allows users to specify optimizations for UDFs with a feature dimension schedule (FDS). {\system} speeds up end-to-end GNN training and inference by up to 32$\times$ on CPU and 7$\times$ on GPU.

\end{abstract}

\section{Introduction}
\label{sec:introduction}

Graph neural networks (GNNs) are gaining popularity in recent years as a promising approach to machine learning on graphs.
Because of the ability to incorporate multi-dimensional features on vertices and edges as well as graph structure information into a joint embedding for downstream tasks, GNNs have shown successful applications in social network mining~\cite{gnn-social-network}, recommender systems~\cite{PinSage}, molecule analysis~\cite{gnn-chemistry}, combinatorial optimization \cite{gnn-combinatorial}, to name a few.

Driven by this trend, specialized software frameworks are emerging to simplify the development and processing of GNN workloads.
These GNN frameworks are typically built on top of existing deep learning systems.
For example, NeuGraph \cite{neugraph} relies on TensorFlow~\cite{tensorflow}; PyTorch geometric (PyG) \cite{PyG} is built upon PyTorch~\cite{pytorch}; DGL \cite{dgl} supports multiple backends.

Unlike traditional neural network workloads that are dominated by dense operations, GNN workloads consist of both dense and sparse operations.
The sparsity comes from the nature of the graph that normally each vertex only connects with a small number of other vertices.
Empirically, sparse operations in a GNN model account for more than 60\% of the total computation time, when both the sparse and dense operations are fully optimized.
While the deep learning systems have benefited from years of development in optimizing dense operations such as convolution and matrix multiplication, they lack flexible support for sparse operations that are essential for high-performance GNN training and inference.
Specifically, the deep learning systems rely on vendor-provided sparse libraries (e.g., MKL \cite{mkl}, cuSPARSE \cite{cusparse}), which offer highly optimized implementations for only a small subset of the kernels required by diverse GNN models.
% but is hard to extend to others.

On the other hand, graph processing systems have been extensively studied in literature~\cite{pregel, x-stream, powergraph, ligra, gunrock, flashgraph}, offering an alternative solution that expresses computations on graphs with a vertex- and/or edge-centric programming paradigm.
As a representative attempt to circumvent the inflexibility of deep learning systems in handling sparse computations, DGL supports offloading the computation kernels in GNNs to existing graph processing systems such as Ligra \cite{ligra} and Gunrock \cite{gunrock}.

\begin{figure}
\centering

\includegraphics[width=0.95\linewidth]{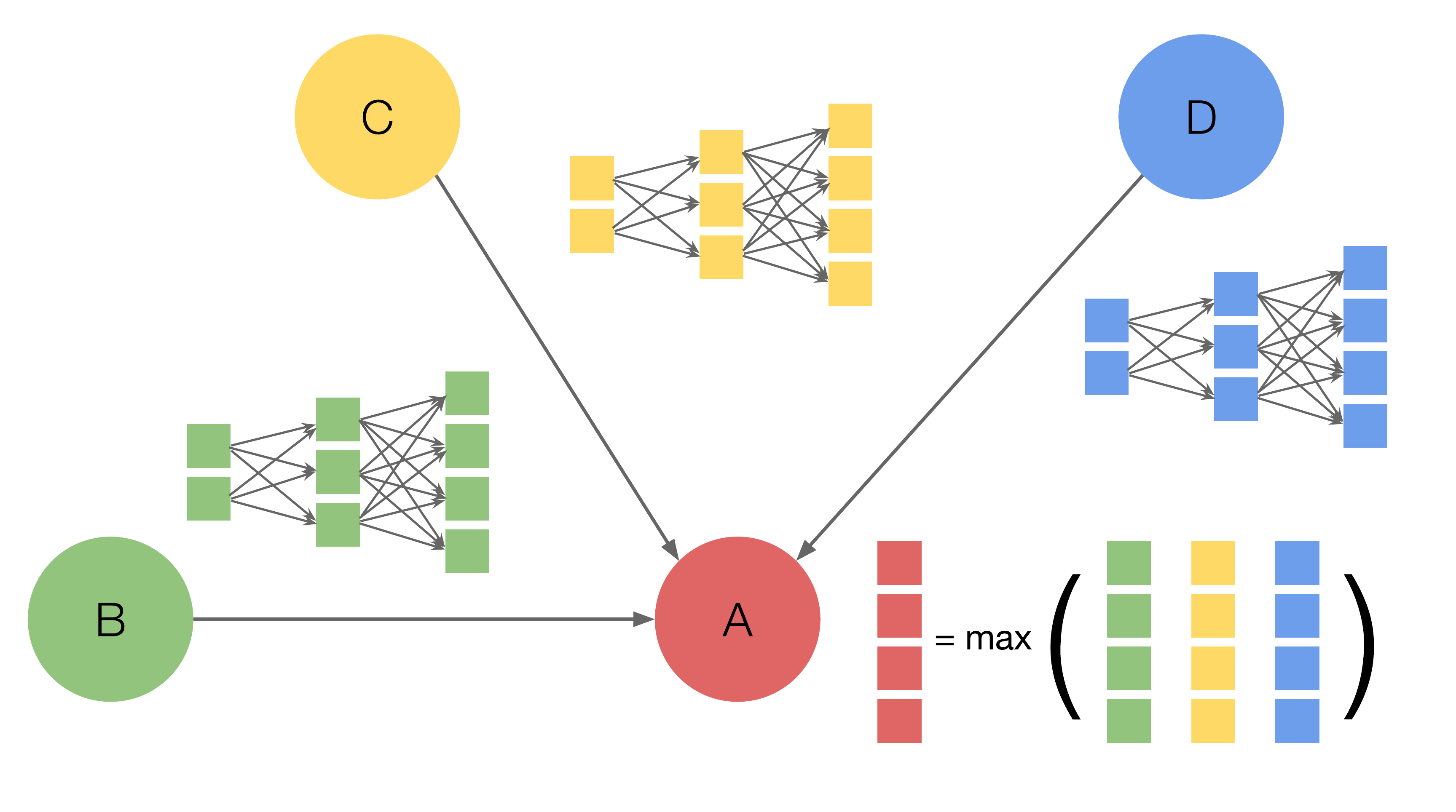}

\caption{Feature dimension computation makes GNNs substantially different from traditional graph workloads --- Shown example is multi-layer perceptron (MLP) aggregation where each edge performs a sequence of vector matrix multiplication and non-linear activation to update the features, and the destination vertex aggregates the features (in this case, picking the maximum) as its representation.}
% \yida{Moving this diagram to page 2 would be better, as it is also mentioned in Section 2}
% \reply{I prefer the first page because I want to convey the message ``Feature dimension computation makes GNNs substantially different from traditional graph workloads" at the very beginning.}
\label{fig:motivation}
\vspace{-1em}
\end{figure}

However, existing graph processing systems are not the panacea, either. They are designed for traditional graph workloads (e.g., BFS, PageRank) where each vertex is associated with a scalar, instead of a feature tensor in GNN's use case.
This additional feature dimension, along with consequently more complex vertex-wise and edge-wise computations, makes kernel optimizations substantially different.
For example, existing graph partitioning techniques aiming at improving cache utilization \cite{cagra, gridgraph} do not take into consideration the feature dimension; hence the entire cache could be occupied by just a few feature tensors.
Also, prior GPU graph processing systems \cite{simd-efficient-graph-gpu, cusha, gunrock} rarely exploit parallelism in feature dimension computation while mainly focusing on designing sophisticated load balancing methods to exploit parallelism in graph traversal.
% \zz{Can we have an example earlier (say, end of Sec 2) to clearly show the limitations of the existing graph processing frameworks? We can move some texts from Sec 3.C}
% \reply{addressed}

% \zd{Given the category of graph processing with linear algebra, I think FeatGraph API is an extension of the linear algebra formulation and is customized for graph neural networks. Most of the graph kernels in graph neural networks are generalized sparse matrix dense matrix multiplication and generalized SDDMM. Thus, we are focusing on optimizing these two sets of graph kernels. We extend the API of GraphBLAS. Instead of accepting binary scalar operators as UDF, we allow more complex operators in UDF.}

% \yida{We didn't talk about the flexibility and expressiveness of the existing solution. We should say something like the existing solutions cannot effectively express the diverse GNN needs, in addition to the sub optimal performance issue.}
% \reply{existing graph processing systems are flexible/expressive, but not efficient for GNN workloads.}
%% insight and design principles %%
This paper proposes \emph{\system} to enable performant processing of GNN workloads.
The key insight is that GNN workloads require co-optimizing graph traversal and feature dimension computation to achieve preferable performance.
\system~suits this need by the design of a two-granularity programming interface.
% \zz{UDF sounds too generic. Can we use a more compelling name here? How about feature dimension schedule (FDS)? We also talk about the mixed sparse-dense nature of GNN workloads which necessitates the co-design of graph traversal and FDS.}
% \reply{FDS is the schedule for UDF. Let's keep UDF since it is a standard name used by graph processing systems.}
% \system~incorporates optimizations for graph traversal into the sparse templates and allows users to specify optimizations for UDFs with a feature dimension schedule (FDS).
% \system~combines these two levels of optimizations to generate high-performance kernels.
% In addition, by decoupling these two levels of optimizations, \system~significantly improves the productivity of developing new kernels for emerging GNN models.
%% design details %%
More concretely, \system~expresses the diverse variants of GNN models by composing \emph{coarse-grained} sparse templates with \emph{fine-grained} feature dimension computations on each vertex/edge in the form of user-defined functions (UDFs).
The coarse-grained level handles traversal over the graph topology, and the fine-grained level focuses on computation over the dense feature tensor of each vertex/edge.
\system~provides two sparse templates: generalized \emph{SpMM} (sparse-dense matrix multiplication) for vertex-wise computations and generalized \emph{SDDMM} (sampled dense-dense matrix multiplication) for edge-wise computations.
% For example, message aggregation on vertices is mapped to the SpMM template, and attention calculation on edges is mapped to the SDDMM template.
Here, \emph{``generalized''} means that the templates can take different fine-grained UDFs.
For example, a commonly used GNN kernel multi-layer perceptron (MLP) aggregation \cite{santoro2017simple, palm2018recurrent}, shown in Figure \ref{fig:motivation}, is mapped to generalized SpMM: it calculates features by performing MLP and aggregates them by taking the max.
% MLP conv is popularly used in a number of GNN models \cite{santoro2017simple, palm2018recurrent}.
Note that the vanilla SpMM operation corresponds to copying features and aggregating them by taking the sum.
Similarly, attention calculation on edges is mapped to generalized SDDMM; the vanilla SDDMM operation corresponds to a specific attention mechanism that performs a dot product between the source vertex feature vector (i.e., 1D tensor) and the destination vertex feature vector.
% \yida{Can these two generalized kernels cover all cases? Also, we should emphasize ``generalized''.}
% \reply{they indeed cover all kernels in GNN's use case as far as I know; Section 3.2 has more discussion on this.}
% \zz{It's not really clear from Sec 3 how our kernels are generalized beyond 2D matrices. Both spmm and sddmm functions take an adjacency matrix as one of the input args. That's it. }\reply{In GNN kernels, the sparse matrix is always the 2D adjacency matrix; the interesting part is that the dense feature tensors can have arbitrary dimensions.}

By cleanly decomposing a kernel specification into sparse templates and UDFs, {\system} enables decoupled, two-level optimizations.
At the coarse-grained level, \system~incorporates optimizations for graph traversal into the sparse templates: applying graph partitioning techniques to improve cache utilization on CPU, adapting parallelization strategies for sparse patterns to fully utilize the massive compute capacity on GPU, etc.
% \yida{Does ``automates'' mean auto-tuning? The aforementioned two techniques are not about automation, and they are not new.}
% \reply{Let's describe it this way: \system~incorporates optimizations for graph traversal into the sparse templates and allows users to specify optimizations for UDFs with a feature dimension schedule (FDS).}
At the fine-grained level, \system~allows users to specify optimizations for UDFs, e.g., how to tile or parallelize feature dimension computation, with a feature dimension schedule (FDS).
\system~combines sparse templates with FDS, and extends a tensor compiler, namely Apache TVM \cite{chen2018tvm}, to generate efficient kernels for both CPU and GPU.
In addition, by decoupling these two levels of optimizations, FeatGraph significantly improves the productivity of developing new kernels for emerging GNN models.

We perform a comprehensive evaluation to verify the efficiency and flexibility of \system. Compared with traditional graph processing systems (i.e., Ligra \cite{ligra} on CPU and Gunrock \cite{gunrock} on GPU), \system~achieves significantly higher performance.
Compared with vendor-provided sparse libraries (i.e., MKL \cite{mkl} on CPU and cuSPARSE \cite{cusparse} on GPU), \system~achieves competitive performance on the kernels that are supported in these libraries while being more flexible to cover more kernels.
We integrated \system~into DGL, a popular GNN framework, to accelerate end-to-end GNN training and inference by up to 32$\times$ on CPU and 7$\times$ on GPU.
To the best of our knowledge, {\system} is the first unified and generic solution that can flexibly integrate with different GNN frameworks and efficiently process GNN workloads on both CPU and GPU.
We summarize the characteristics of \system~and other works in Table~\ref{tab:comparison}.
\final{{\system} is available in open-source format at \url{https://github.com/dglai/FeatGraph}.}

\begin{table}
\centering
\begin{adjustbox}{width=\linewidth}
\begin{tabular}{l|cccc}
\toprule
& Platform & Flexibility & Efficiency & Open-source \\
\midrule
MKL \cite{mkl}            & CPU              & low  & high & no  \\
cuSPARSE \cite{cusparse}  & GPU              & low  & high & no  \\
Ligra \cite{ligra}        & CPU              & high & low  & yes \\
Gunrock \cite{gunrock}    & GPU              & high & low  & yes \\
{\system}                 & CPU and GPU      & high & high & yes \\
\bottomrule
\end{tabular}
\end{adjustbox}
\caption{Side-by-side comparison between {\system} and existing works on handling GNN workloads.}
\label{tab:comparison}
\vspace{-1em}
\end{table}

Specifically, this paper makes the following contributions:
\begin{itemize}
    \itemsep0cm
    \item {\system} provides a flexible programming interface that is able to express the diverse variants of GNN models by composing coarse-grained sparse templates with customizable fine-grained feature dimension computations on each vertex/edge.
    \item {\system} performs extensive optimizations in both graph traversal and feature dimension computation to generate efficient kernels. In addition, {\system} decouples these two levels of optimizations to improve the productivity of developing new kernels for emerging GNN models.
    \item Experiment results on representative GNN models and a wide collection of datasets show that \system~is portable to existing GNN frameworks and serves as a flexible and efficient backend.
    % \zd{in DGL?} \reply{representative models are not necessarily connected to DGL framework}
\end{itemize}

The rest of the paper is organized as follows.
Section \ref{sec:background} reviews the background of GNNs and tensor compilers, and motivates {\system} by examining the limitations of existing graph processing systems.
Section \ref{sec:design} describes the programming interface design and optimization techniques of {\system}.
Section \ref{sec:implementation} presents the system implementation, followed by evaluation in Section \ref{sec:evaluation}.
We discuss related work in Section \ref{sec:related} and summarize in Section \ref{sec:conclusion}.

\section{Background and Motivation}
\label{sec:background}
%%%%%%%%%%%%%%%%%
% Notation rules:
%   - \mathcal for graph, set of vertices, set of edges
%   - \mathbf for matrices and vectors
%   - \mathtt for function names
%   - index is in lower case, non-bold
%%%%%%%%%%%%%%%%%

\subsection{Graph Neural Networks (GNNs)}
\label{subsec:gnn}

In recent years, there is a rise of interest in adopting deep learning to structural data such as graphs. Unlike the dense objects (e.g., images, videos, texts) handled by traditional deep learning models, graphs represent sparsely, irregularly connected links. Essentially, graphs are defined on a non-Euclidean domain equipped with vastly different distance measurements and geometric properties, imposing the demand for new neural network architectures.

GNNs are an emerging family of neural networks capable of learning a joint representation for each vertex/edge using both features and topological data.
Recent studies~\cite{gnn-chemistry, graphnets} have unified different GNN models into a \emph{message passing} paradigm where each vertex computes a new representation by aggregating features (messages) from its neighbors.
More formally, given a graph $\mathcal{G(\mathcal{V},\mathcal{E})}$, we denote the input feature tensor associated with vertex $v$ as $\mathbf{x}_v$, and that associated with the edge pointing from vertex $u$ to $v$ as $\mathbf{x}_{uv}$.
To get the representation of a vertex and an edge, the message passing paradigm carries out the following computations:

% \vspace{-1em}

\begin{equation}\label{eq:mp-vertex}
\mathbf{h}_v = \bigoplus_{u\in\mathcal{N}(v)}\phi(\mathbf{x}_u, \mathbf{x}_v, \mathbf{x}_{uv})
\end{equation}

% \vspace{-0.5em}

\begin{equation}\label{eq:mp-edge}
\mathbf{h}_{uv} = \psi(\mathbf{x}_u, \mathbf{x}_v, \mathbf{x}_{uv})
\end{equation}

% \vspace{-1em}

Here $\phi$, $\bigoplus$, and $\psi$ are customizable or parameterized functions (e.g., neural network modules) for calculating messages, aggregating messages, and updating edge representations, respectively.
Similar to convolutional neural networks (CNNs), a GNN model iteratively applies Equations~\eqref{eq:mp-vertex}~\eqref{eq:mp-edge} to generate vertex and edge representations for higher layers.

There is a strong connection between Equations~\eqref{eq:mp-vertex}~\eqref{eq:mp-edge} and sparse matrix operations.
For example, given the vertex feature matrix $\mathbf{X_V}\in \mathbb{R}^{|\mathcal{V}|\times d}$ and the adjacency matrix $\mathbf{A}$, the vertex-wise computation in the graph convolutional network (GCN)~\cite{gcn}, which copies source vertex features as messages and aggregates messages by taking the sum, is equivalent to SpMM (sparse-dense matrix multiplication) as follows.

\begin{equation}\label{eq:spmm}
\mathbf{H_V}=\mathbf{A}\times \mathbf{X_V}
\end{equation}

For edge-wise computations, many GNN models~\cite{gat, agnn} calculate an attention weight on each edge.
One popular formulation for calculating attention weight is by a dot product between the source and destination vertex features~\cite{vaswani2017attention}, that is, $\psi(\mathbf{x}_u, \mathbf{x}_v, \mathbf{x}_{uv})\triangleq \mathbf{x}_u\mathbf{x}_v^T$.
Its tensorized implementation corresponds to SDDMM (sampled dense-dense matrix multiplication) \cite{sddmm-2}, which multiplies two dense matrices, followed by an element-wise multiplication with a sparse mask matrix, to output a sparse matrix.

\begin{equation}\label{eq:sddmm}
\mathbf{H_E}=\mathbf{A}\cdot (\mathbf{X_V}\times \mathbf{X_V}^T)
\end{equation}

Hence, Equations~\eqref{eq:mp-vertex} and~\eqref{eq:mp-edge} when implemented as tensor operations are generalized SpMM and SDDMM, respectively.
They represent two distinct computation patterns in GNN workloads: reduction on each vertex and reduction on each edge.
Moreover, according to the chain rule, the gradient computation of SpMM with respect to $\mathbf{A}$ requires a dot product between the gradients of source and destination vertex features, thus following the SDDMM pattern.
Likewise, the gradient computation of SDDMM follows the SpMM pattern.
Therefore, these two computation patterns are essential for both inference and training of GNNs.
In particular, our benchmarking shows that generalized SpMM and SDDMM occupy $\sim95\%$ of the total run time in training a 2-layer GNN model, using the existing solutions with sub-optimized sparse kernels.

\subsection{Limitations of Existing Graph Processing Systems}
Existing graph processing systems \cite{ligra, gunrock} express computations on graphs with a vertex- and/or edge-centric programming paradigm, and they employ a scheduler to realize efficient graph traversal.
For example, to ensure load balance on GPU, Gunrock \cite{gunrock} assigns the edges of a vertex to be processed by a thread, a warp, or a block, according to the number of the edges.
Edge is the unit for scheduling---the computation on an edge is blackbox to the scheduler.
The underlying assumption is that the computation on an edge is lightweight.

However, that assumption breaks in GNNs, which attach a multi-dimensional feature tensor to each vertex/edge, and consequently have more complex vertex-wise and edge-wise computations than traditional graph workloads.
For example, MLP aggregation, as shown in Figure \ref{fig:motivation}, performs a sequence of vector matrix multiplication and non-linear activation on each edge.
Treating the computation on each edge as a blackbox, Gunrock fails to exploit the abundant parallelism in it.

To enable performant processing of GNN workloads, we need a system that: 1) makes vertex-wise and edge-wise computations whitebox to the scheduler; 2) co-optimizes graph traversal and feature dimension computation.
{\system} achieves the goals by adopting a tensor compiler approach.

\subsection{Tensor Compiler}
Computation-intensive workloads typically operate on tensors, i.e., multi-dimensional data.
For example, traditional deep learning models perform matrix multiplication and convolution over dense tensors; GNNs deal with both dense and sparse tensors (the feature tensor is dense and the adjacency matrix is sparse).
Previously, people rely on vendor-specific libraries (e.g., MKL, cuBLAS) to obtain high performance of tensor computations over the vendors' own CPUs and GPUs.
These libraries require heavy manual tuning by experienced engineers.
As a result, they evolve slowly in contrast with the rapid emergence of new workloads.

An alternative solution is tensor \revision{compilation}, which expresses the processing of tensors in its own intermediate representation (IR) \cite{ragan2013halide, chen2018tvm}.
Tensor compilation separates the computation definition (i.e., what to compute) from the scheduling (i.e., how to compute) so as to focus on the scheduling part for performance optimization.
A scheduling scheme can apply loop transformations, vectorization, thread binding, etc., to manipulate the tensor computation.
Optimizing one computation kernel for different hardware architectures is essentially searching for different scheduling schemes.

\system~adopts the tensor \revision{compilation} approach to optimize the computation kernels in GNN workloads.
However, existing tensor compilers \cite{chen2018tvm, ragan2013halide} mostly focus on computations over dense tensors, and there is little support for computations involving sparse tensors.
\system~extends TVM~\cite{chen2018tvm} to support the core sparse patterns in GNNs (i.e., generalized SpMM and SDDMM), and allows customizable feature dimension computations on each vertex/edge by the design of a two-granularity programming interface (Sec \ref{subsec:api}).

\section{System Design and Optimization}
\label{sec:design}

In this section, we first give an overview of the software stack of \system~(Sec~\ref{subsec:overview}).
We then describe the design of the programming interface and demonstrate its expressiveness using code examples (Sec~\ref{subsec:api}).
Finally, we cover the optimization techniques for generating efficient GNN kernels on CPU and GPU (Sec~\ref{subsec:opt}).

\begin{figure}
\centering

\includegraphics[width=0.95\linewidth]{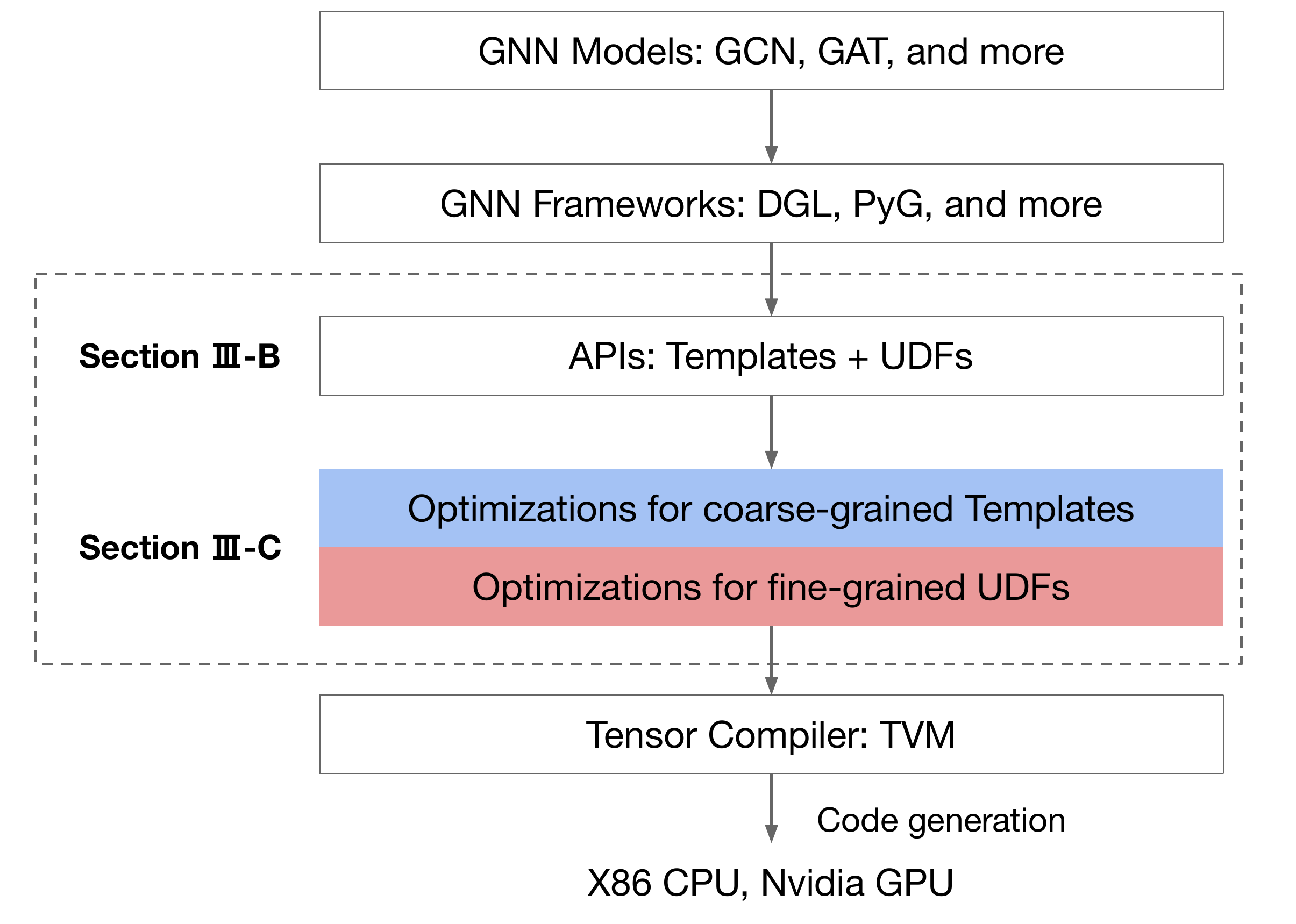}

\caption{System overview of \system.}
\label{fig:system-overview}
\vspace{-1em}
\end{figure}

\subsection{System Overview}
\label{subsec:overview}

Figure \ref{fig:system-overview} depicts the software stack of \system.
At the top level, users define, train, and evaluate GNN models in specialized frameworks such as DGL and PyG, \revision{which handle dataflow programming and automatic differentiation.}
\system~serves as a backend for these frameworks, targeting the message passing computation that is core to GNN workloads.
\system~provides a flexible programming interface to express the diverse variants allowed by the message passing paradigm.
Specifically, \system~describes feature dimension computations on each vertex/edge with user-defined functions (UDFs), and triggers UDFs by SpMM or SDDMM sparse template.
{\system} incorporates optimizations for graph traversal into the sparse templates, and allows users to specify optimizations for UDFs with a feature dimension schedule (FDS).
\system~combines templates with FDS, and leverages the TVM tensor compiler \cite{chen2018tvm} to generate efficient kernels for both CPU and GPU.
By decoupling these two levels of optimizations, \system~significantly improves the productivity of developing new kernels for emerging GNN models.

\subsection{Programming Interface}

\label{subsec:api}

\begin{figure}
\hspace{0.5em}
\begin{subfigure}{\linewidth}
\begin{lstlisting}[language=python]
import featgraph, tvm
A = featgraph.spmat(shape=(n,n), nnz=m)

# use src vertex feature as message
XV = tvm.placeholder(shape=(n,d))
def msgfunc(src, dst, eid):
  out = tvm.compute((d,), lambda i: XV[src,i])
  return out

# tile feature dimension for cache optimization
def cpu_schedule(out):
  s = tvm.create_schedule(out)
  # the tiling factor is tunable
  s[out].split(out.axis[0], factor=8)
  return s

# parallelize feature dimension
# by binding it to the thread index in CUDA
def gpu_schedule(out):
  s = tvm.create_schedule(out)
  s[out].bind(out.axis[0], 'thread.x')
  return s

# use sum as the aggregation function
aggregation = tvm.sum

# trigger the SpMM template
if target = 'cpu':
  fds = cpu_schedule
elif target == 'gpu':
  fds = gpu_schedule
GCN = [*\bfseries\color{blue}featgraph.spmm*](A, msgfunc, aggregation,
                     target, [*\bfseries\color{red}fds*])
\end{lstlisting}
\vspace{-0.8em}
\caption{GCN aggregation}
\label{fig:spmm-code-gcn-conv}
\end{subfigure}

\hspace{0.5em}
\begin{subfigure}{\linewidth}
\begin{lstlisting}[language=python]
# ReLU((src feature + dst feature) * W)
XV = tvm.placeholder(shape=(n,d1))
W = tvm.placeholder(shape=(d1,d2))
def msgfunc(src, dst, eid):
  k = tvm.reduce_axis((0,d1))
  out = tvm.compute((d2,), lambda i:
    tvm.max(tvm.sum((XV[src,k] + XV[dst,k])
                     * W[k,i])), 0)
  return out

\end{lstlisting}
\vspace{-0.8em}
\caption{Message function of MLP aggregation}
\label{fig:spmm-code-mlp-conv}
\end{subfigure}

\caption{Example code of vertex-wise computations with the SpMM template --- {\system} inlines the fine-grained FDS (in red) into the coarse-grained SpMM template (in blue) to generate a fused, optimized kernel.}
%\caption{Vertex-wise computations with the SpMM template --- {\system} incorporates optimizations for graph traversal into the SpMM template and allows users to specify optimizations for the message function with an FDS. After performing these two levels of optimizations, {\system} inlines the message function into the SpMM template to generate a fused kernel.}
\label{fig:spmm-code}
\vspace{-0.5em}
\end{figure}

\begin{figure}

\begin{subfigure}{\linewidth}
\begin{lstlisting}[language=python]
import featgraph, tvm
A = featgraph.spmat(shape=(n,n), nnz=m)

# dot product between src and dst vertex features
XV = tvm.placeholder(shape=(n,d))
def edgefunc(src, dst, eid):
  k = tvm.reduce_axis((0,d))
  out = tvm.compute(shape=(1,), lambda i:
	  tvm.sum(XV[src,k] * XV[dst,k]))
  return out

# tree-based parallel reduction
def gpu_schedule(out):
  s = tvm.create_schedule(out)
  s[out].tree_reduce(out.reduce_axis[0], 'thread.x')
  return s

# trigger the SDDMM template
target = 'gpu'
fds = gpu_schedule
Attention = [*\bfseries\color{blue} featgraph.sddmm*](A, edgefunc, target, [*\bfseries\color{red} fds*])
\end{lstlisting}
\vspace{-0.8em}
\caption{Dot-product attention}
% \yida{why only GPU scheduling}
% \reply{CPU schedule is the same as in GCN conv; save some space}
\label{fig:sddmm-code-dot-product-attention}
\end{subfigure}

\begin{subfigure}{\linewidth}
\begin{lstlisting}[language=python]
# multiple dot products
XV = tvm.placeholder(shape=(n,h,d))
def edgefunc(src, dst, eid):
  k = tvm.reduce_axis((0,d))
  out = tvm.compute(shape=(h,), lambda i:
    tvm.sum(XV[src,i,k] * XV[dst,i,k]))
  return out
\end{lstlisting}
\vspace{-0.8em}
\caption{Edge function of multi-head dot-product attention}
\label{fig:sddmm-code-multi-head-dot-product-attention}
\end{subfigure}

\caption{Example code of edge-wise computations with the SDDMM template --- {\system} inlines the fine-grained FDS (in red) into the coarse-grained SDDMM template (in blue) to generate a fused, optimized kernel.}
\label{fig:sddmm-code}
\vspace{-0.5em}
\end{figure}

%% design principles %%
There are two principles in the design of {\system}'s programming interface.
First, the interface should closely follow the mathematical definition of GNNs as described in Section \ref{subsec:gnn}.
Second, it should facilitate optimizations.

%% high level description %%
To these ends, we propose to decompose a kernel specification into two parts: UDFs written in a tensor expression language adopted from TVM to describe fine-grained feature dimension computations on each vertex/edge, and the choice of coarse-grained sparse patterns.
\system~provides two kernel templates \texttt{featgraph.spmm} and \texttt{featgraph.sddmm} for the SpMM and SDDMM sparse patterns that directly map to the vertex-wise and edge-wise computations in the message passing paradigm, i.e., Equations \eqref{eq:mp-vertex} and \eqref{eq:mp-edge}.

%% detailed description of spmm %%
More concretely, \texttt{featgraph.spmm} takes in five arguments: an adjacency matrix, a message function, an aggregation function, the target (CPU or GPU), and an FDS to specify optimizations of the message function.
Figure~\ref{fig:spmm-code-gcn-conv} shows the code for GCN aggregation, i.e., the message aggregation in GCN model as described in Section \ref{subsec:gnn}.
% \yida{Did we define GCN conv before?}.
% \reply{resolved}
Given the edge ID tuple \texttt{(src, dst, eid)}, the user-defined message function \texttt{msgfunc} (line 6--8) slices out the \texttt{src} row from the vertex feature matrix \texttt{XV}, which is equivalent to using the source vertex feature as the message.
The aggregation function is \texttt{sum} and any commutative reducer is allowed.
% The same API can express more complex cases with different definitions of computation (i.e., the message function) and scheduling (i.e., the FDS).
Figure~\ref{fig:spmm-code-mlp-conv} shows a more complex message function, which adds the source and destination vertex features, and then multiplies with a weight matrix, followed by a ReLU activation (i.e., taking the max with 0).
% We will talk about the FDS for MLP conv in Section~\ref{subsec:opt}.

%% some discussion; fusing the message function %%
{\system} can easily support the commonly used message functions in GNNs---specifically, all the builtin ones provided by DGL\footnote{\url{https://docs.dgl.ai/api/python/function.html\#message-functions}}, including copying vertex or edge feature tensor, element-wise operations between vertex and edge feature tensors, etc.
In addition, {\system} can express more complex message functions such as the one in MLP aggregation.

{\system} inlines the message function into the SpMM template to generate a fused kernel.
In contrast, existing GNN frameworks (e.g., DGL, PyG, NeuGraph) that rely on deep learning systems as backend have to materialize the messages on every edge, causing inefficiency in both performance and memory consumption. 

%% detailed description of sddmm %%
\texttt{featgraph.sddmm} takes in four arguments: an adjacency matrix, an edge function, the target (CPU or GPU), and an FDS to specify optimizations of the edge function.
Figure~\ref{fig:sddmm-code-dot-product-attention} shows the code for dot-product attention, where the user-defined edge function \texttt{edgefunc} (line 6--10) performs a dot product between the source and destination vertex feature vectors, and returns an attention weight as the new feature on the edge.
% When the edge function is more complex, we just need to define the compute and schedule of it accordingly, but still use the same APIs.
Figure~\ref{fig:sddmm-code-multi-head-dot-product-attention} shows a more complex edge function, which performs multiple dot products over the feature tensors.
% Again, the FDS of this edge function will be discussed in Section~\ref{subsec:opt}.

%% schedule for UDFs %%
This two-granularity programming interface simplifies implementing new GNN kernels and, more important, facilitates optimizations.
By cleanly decomposing a kernel specification into coarse-grained sparse templates and fine-grained feature dimension computations on each vertex/edge in the form of UDFs, {\system} enables decoupled, two-level optimizations.
Specifically, {\system} incorporates optimizations for graph traversal into the sparse templates and allows users to specify optimizations for UDFs with an FDS.
Some FDS examples, both for CPU and for GPU, are shown in Figure~\ref{fig:spmm-code-gcn-conv} at line 11--15 and line 19--22.
It is worth noting that when the FDS is missing, \system~essentially degrades to traditional graph processing systems that are designed without special handling of feature dimension computation.

\begin{figure}[b]
\centering

% we can't have blank lines in between, otherwise the two subfigures will be stacked
% instead of being placed side by side
\begin{subfigure}[c]{0.3\linewidth}
\includegraphics[width=1.0\linewidth]{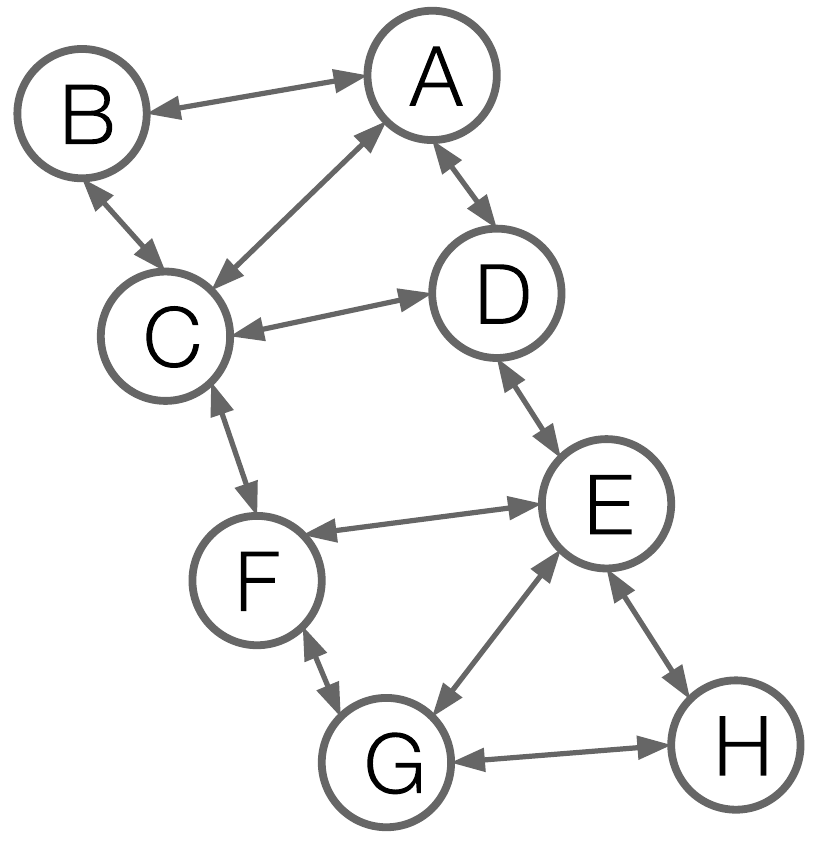}
\label{fig:sample-graph-sub1}
\end{subfigure}
\begin{subfigure}[c]{0.45\linewidth}
\includegraphics[width=1.0\linewidth]{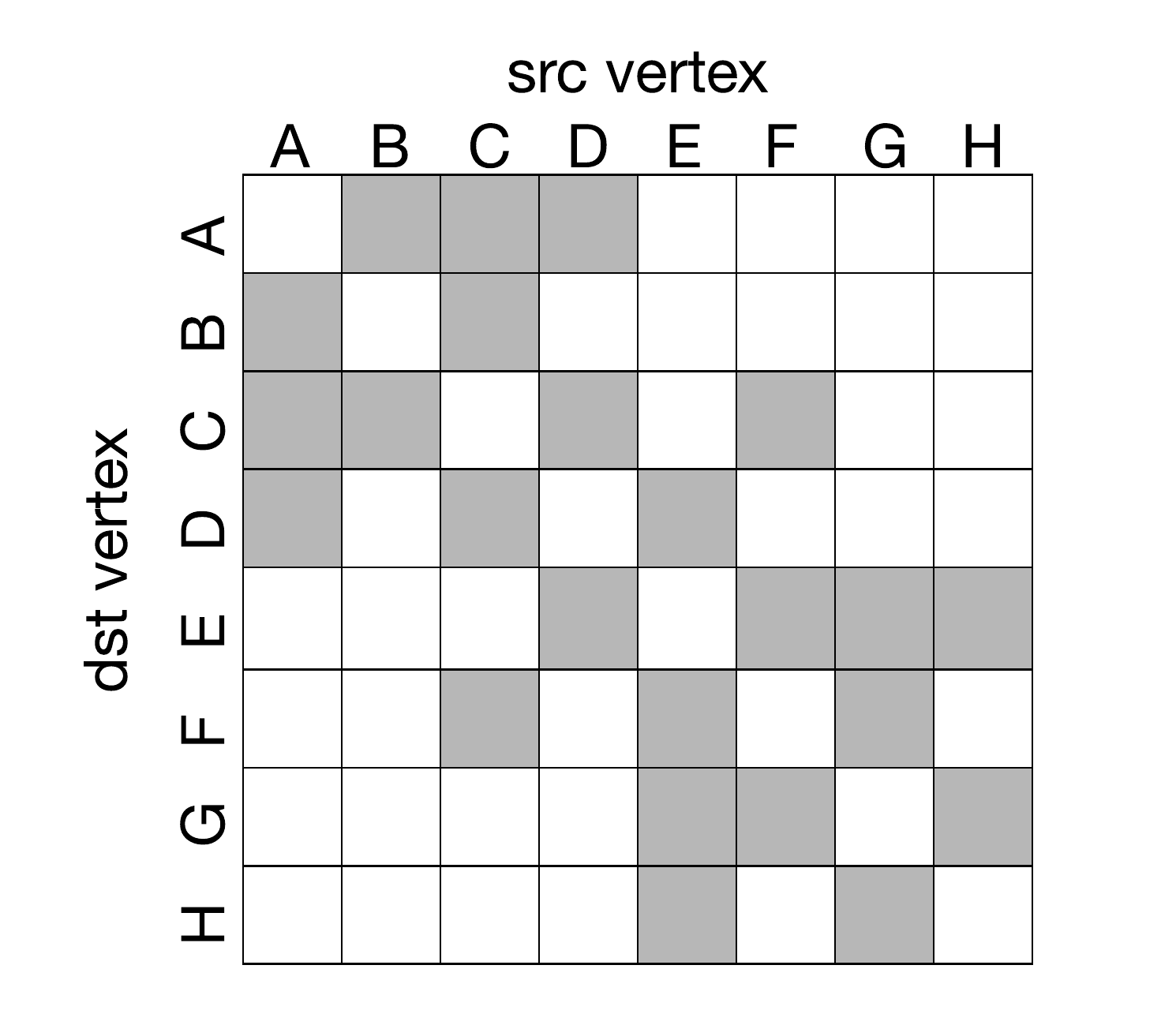}
\label{fig:sample-graph-sub2}
\end{subfigure}

\caption{A sample graph with 8 vertices and its corresponding adjacency matrix.}
\label{fig:sample-graph}
\vspace{-1em}
\end{figure}

\begin{figure*}[t]
\centering

\begin{subfigure}[b]{0.35\linewidth}
\includegraphics[width=0.85\linewidth]{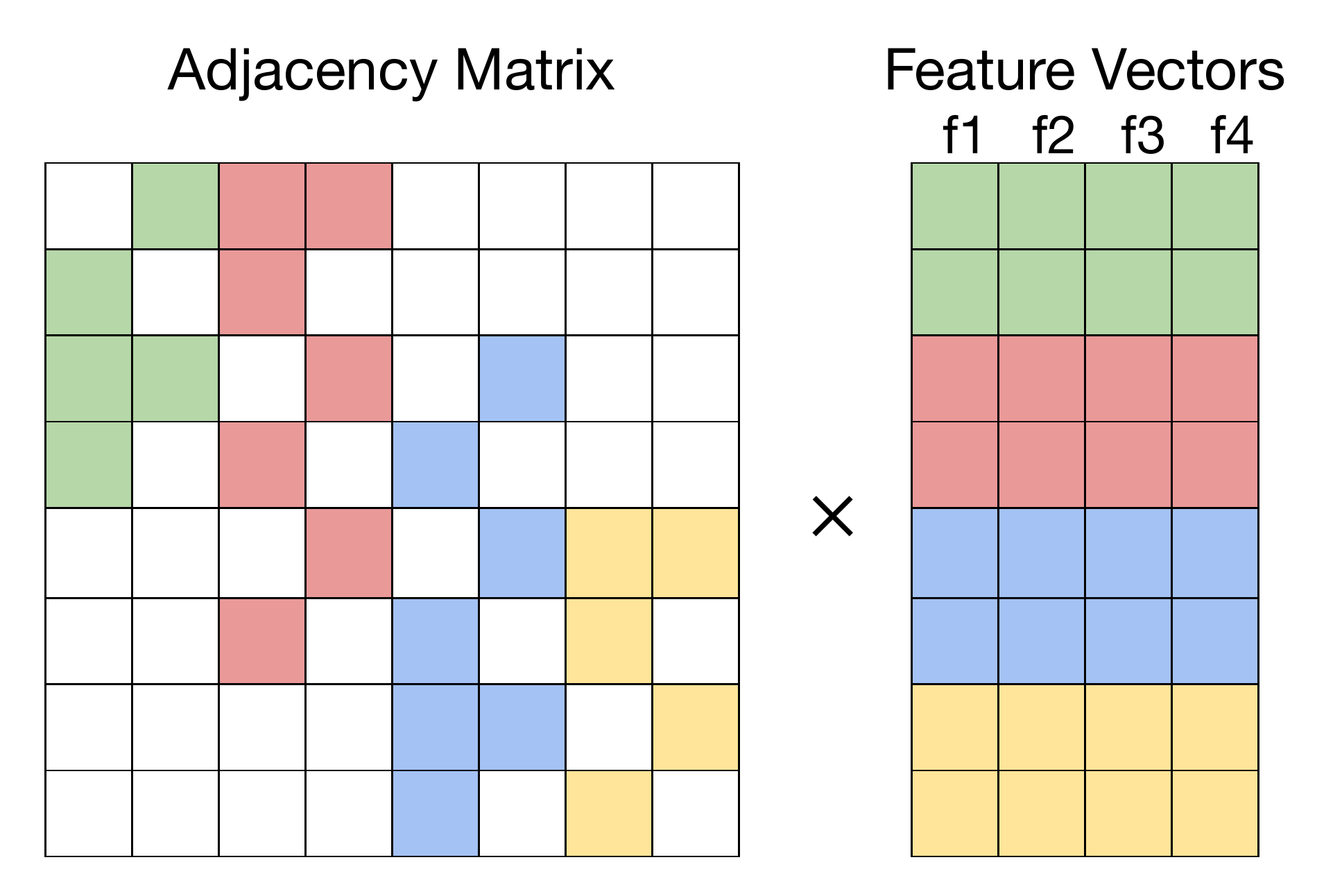}
\caption{1D graph partitioning}
\label{fig:SpMM-partition-sub1}
\end{subfigure}
\begin{subfigure}[b]{0.55\linewidth}
\includegraphics[width=1.0\linewidth]{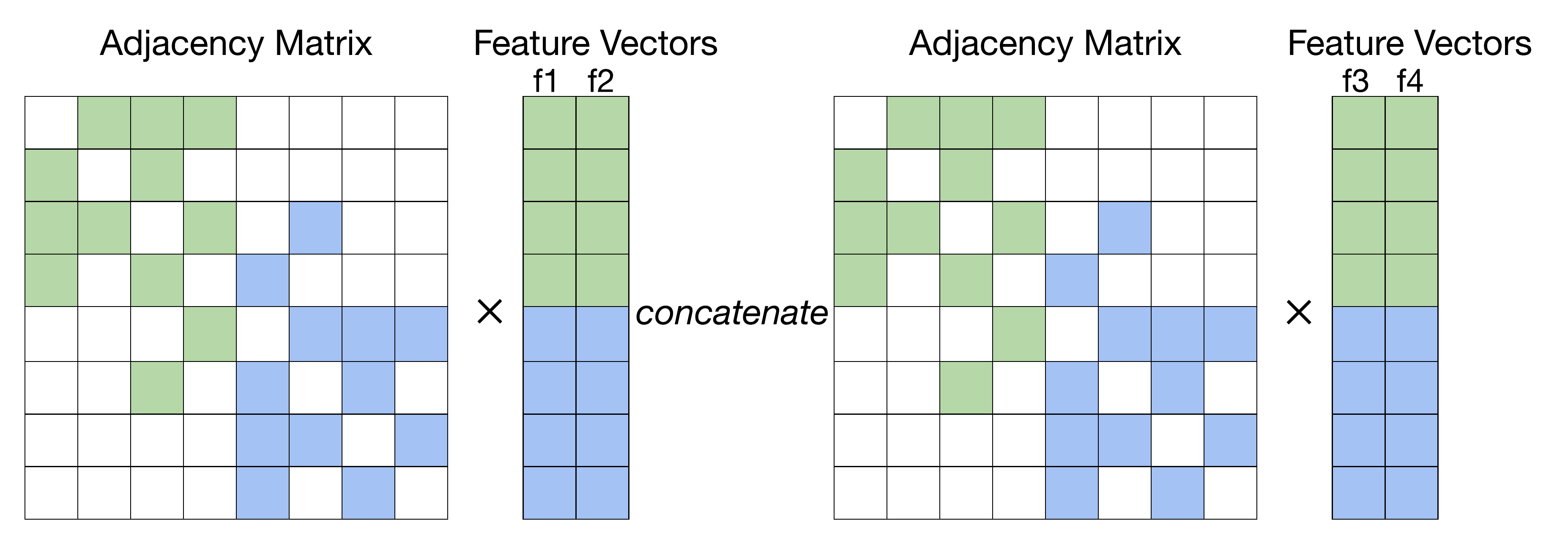}
\caption{1D graph partitioning combined with feature dimension tiling}
\label{fig:SpMM-partition-sub2}
\end{subfigure}

\caption{Feature dimension tiling and 1D graph partitioning for cache optimization in GCN aggregation --- We assume the cache can hold two feature vectors. Tiling each feature vector into two sub-vectors reduces the number of graph partitions from four to two, which translates to 50\% saving in merge, but at the cost of accessing the adjacency matrix twice.}
\label{fig:SpMM-partition}
\vspace{-1.0em}
\end{figure*}

\begin{figure*}[t]
\centering

\begin{subfigure}[b]{0.4\linewidth}
\includegraphics[width=0.85\linewidth]{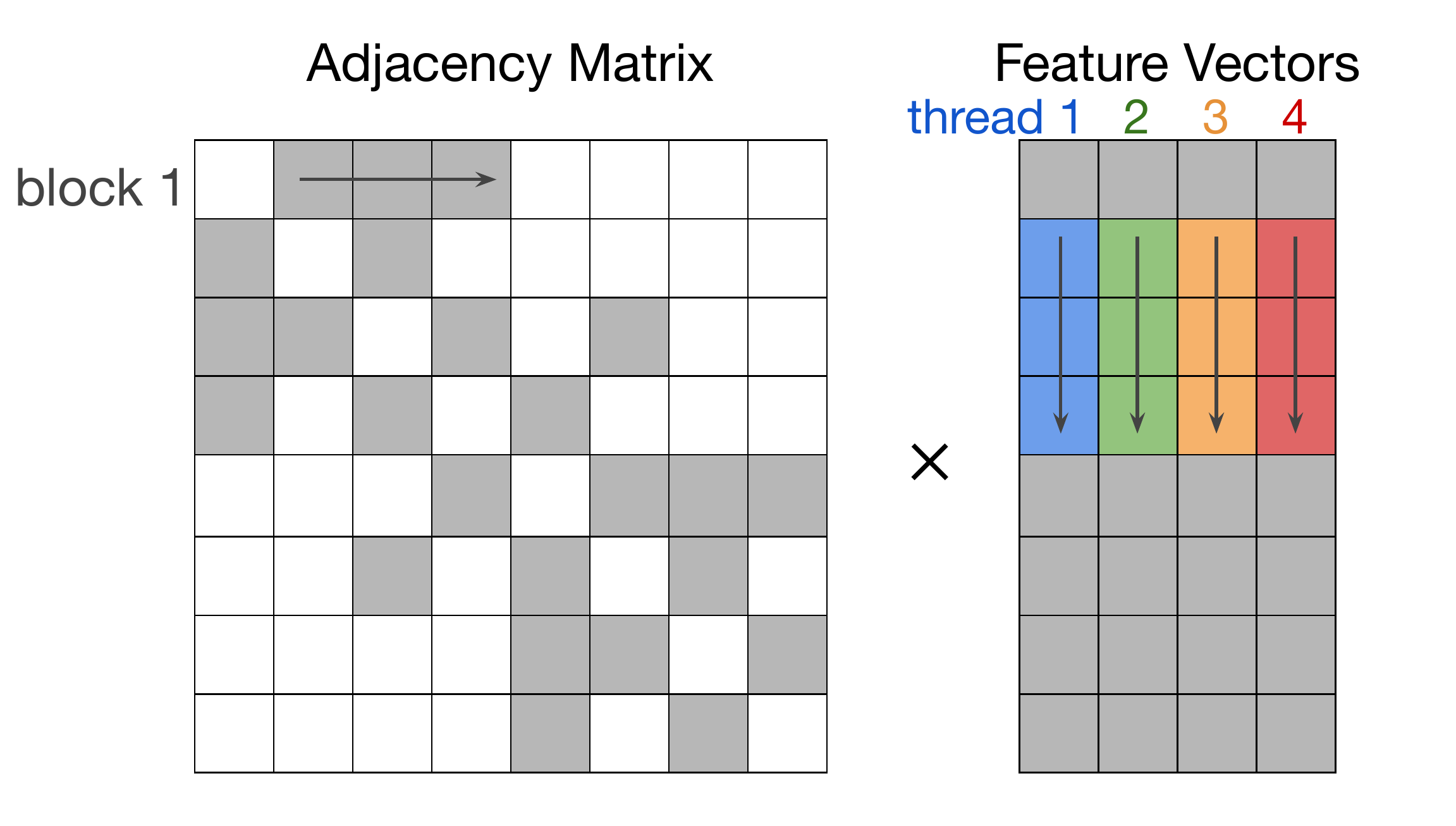}
\vspace{10pt}
\caption{GCN aggregation}
\label{fig:specialized-parallelization-gpu-sub1}
\end{subfigure}
\begin{subfigure}[b]{0.58\linewidth}
\includegraphics[width=1.0\linewidth]{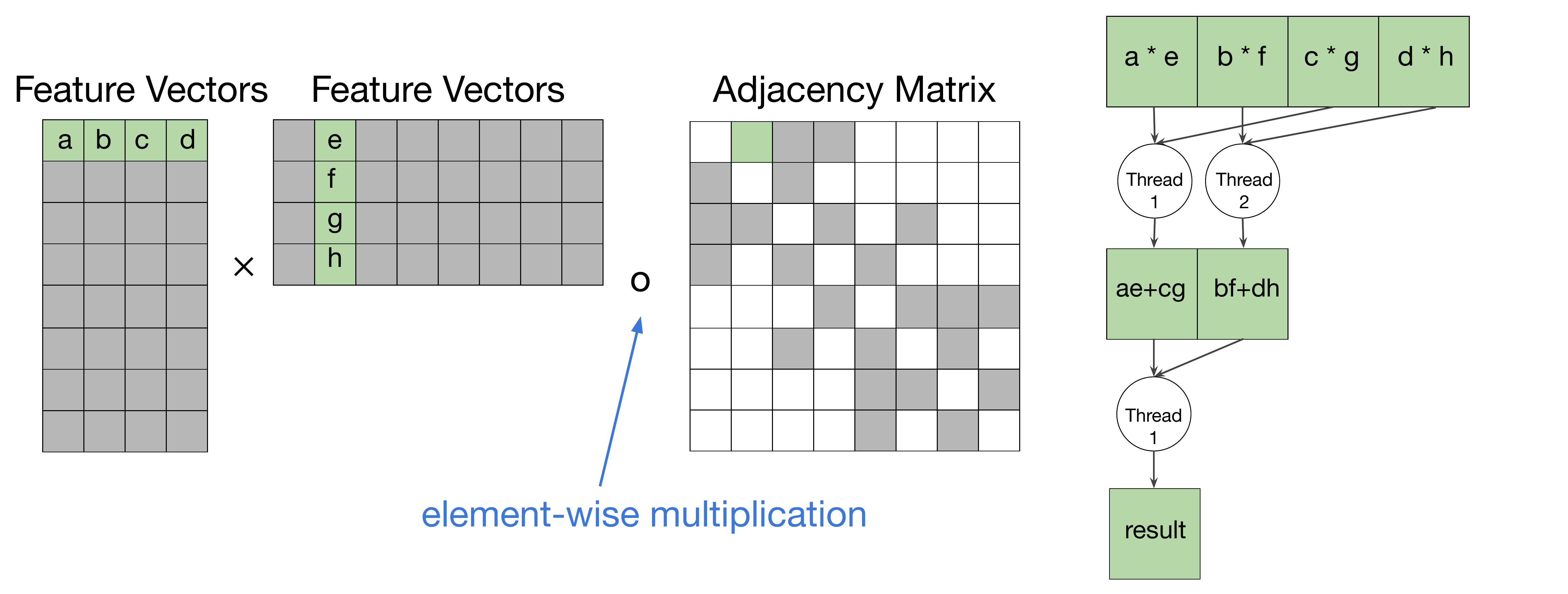}
\caption{Dot-product attention}
\label{fig:specialized-parallelization-gpu-sub2}
\end{subfigure}

\caption{Parallelization strategies adapted for computation patterns.}
\label{fig:specialized-parallelization-gpu}
\vspace{-0.5em}
\end{figure*}

\subsection{Decoupled, Two-level Optimizations}
\label{subsec:opt}

% \zz{We need to revisit and elaborate our novelty claim on co-optimizing traversal and feature dimension computation.}
% \reply{addressed}

This subsection describes the optimizations for graph traversal, which are incorporated into the sparse templates, and the optimizations for feature dimension computation, which are specified by users with an FDS.
We analyze the interplay between these two levels of optimizations, and show that by combining them, {\system} enables performant processing of GNN workloads. 
Throughout this subsection, we use the sample graph shown in Figure \ref{fig:sample-graph} to illustrate optimization techniques.

\subsubsection{Graph Partitioning and Feature Dimension Tiling}

On CPU, the key factor limiting the efficiency of graph traversal is poor locality, which causes low cache utilization.
Prior arts have attempted to improve locality in graph traversal by graph partitioning \cite{cagra, gridgraph}.
{\system} proposes combining graph partitioning with feature dimension tiling to strike a balance between efficiency of graph traversal and efficiency of feature dimension computation.

Figure \ref{fig:SpMM-partition} illustrates how feature dimension tiling is combined with 1D graph partitioning \cite{cagra}, which partitions source vertices, to effectively optimize cache utilization in GCN aggregation, i.e., the vanilla SpMM operation.
Here we assume the feature vector length is four, and the cache can hold two feature vectors.
With 1D graph partitioning alone, source vertices are partitioned into four segments so that each segment fits into the cache; these segments are processed one by one to get four portions of intermediate results; in the end the intermediate results are merged.
1D graph partitioning improves read locality within each segment at the cost of merging intermediate results from different segments.
When 1D graph partitioning is combined with feature dimension tiling, merge cost is reduced since more vertices can fit into the cache under the same capacity.
As shown in Figure \ref{fig:SpMM-partition-sub2}, tiling each feature vector into two sub-vectors reduces the number of segments from four to two, which translates to 50\% saving in merge cost.
However, feature dimension tiling results in traversing the graph twice, which means increased accesses to graph topological data (i.e., the adjacency matrix).

Thus, feature dimension tiling introduces the trade-off between accesses to graph topological data and accesses to feature data.
In GNNs, feature vectors have a typical length ranging from 32 to 1024.
When the tiling factor is properly selected, the gain of improved locality in accessing feature data far outweighs the overhead of increased accesses to graph topological data.

More complex UDFs that compute on multi-dimensional feature tensors may require a multi-level tiling scheme.
To efficiently support diverse UDFs, {\system} allows users to specify optimizations for UDFs with an FDS.
Figure \ref{fig:schedule-mlp-conv-cpu} shows the FDS for MLP aggregation---it tiles both dimensions of the weight matrix for cache optimization.

For edge-wise computations, besides feature dimension tiling, {\system} employs a graph traversal scheme \cite{hilbert} based on Hilbert curve.
% \yida{The introduction of Hilbert curve traversal is too abrupt. Why we need it, solving what problem, what does it do? We should give some setup.}
Edge-wise computations access both source and destination vertex features, and update edge features; Hilbert curve traversal exploits locality in accessing both source and destination vertices.
The recursive structure of Hilbert curve enables exploiting locality across a spectrum of granularities, e.g., L1/L2/L3 caches.
% However, Hilbert curve traversal is hard to parallelize---parallelization destroys locality.
% To solve this issue, we propose coarsened Hilbert curve traversal that partitions the adjacency matrix into blocks, traverses non-zero elements (which correspond to edges) within each block in a normal order (i.e., row by row), and traverses different blocks in Hilbert curve order.
% This coarsened Hilbert curve traversal scheme permits intra-block parallelization while preserving inter-block locality.
{\system} combines Hilbert curve traversal with feature dimension tiling to fully optimize edge-wise computations.

\begin{figure}
\begin{lstlisting}[language=python]
# tile multiple dimensions
def cpu_schedule(out):
  s = tvm.create_schedule(out)
  s[out].split(out.axis[0], factor=8)
  s[out].split(out.reduce_axis[0], factor=8)
  return s
\end{lstlisting}
\vspace{-1em}
\caption{FDS for MLP aggregation on CPU.}
\vspace{-1em}
\label{fig:schedule-mlp-conv-cpu}
\end{figure}

\begin{figure}
\begin{lstlisting}[language=python]
# parallelize multiple dimensions
def gpu_schedule(out):
  s = tvm.create_schedule(out)
  s[out].bind(out.axis[0], 'block.x')
  s[out].tree_reduce(out.reduce_axis[0], 'thread.x')
  return s
\end{lstlisting}
\vspace{-1em}
\caption{FDS for MLP aggregation on GPU.}
\vspace{-1em}
\label{fig:schedule-mlp-conv-gpu}
\end{figure}

\subsubsection{Adaptive Parallelization Strategies}
% \zz{I suggest we explicitly explain what is adaptive in this scheme}
% \reply{addressed}

To utilize GPU's massive parallel compute capacity, prior graph processing systems exploit parallelism in graph traversal by implementing either vertex parallelization or edge parallelization \cite{gunrock, cusha, simd-efficient-graph-gpu}.
However, they are unable to exploit the abundant parallelism in feature dimension computation arising in GNN workloads due to treating the UDFs as a blackbox.
{\system} enables exploiting parallelism in feature dimension computation by opening the blackbox of UDFs so as to inform the scheduler.
Specifically, {\system} allows users to specify a parallelization scheme for UDFs with an FDS, which can be adapted to the diverse computation patterns of UDFs to fully exploit the parallelism in feature dimension computation.

% In this work, we focus on exploiting parallelism in feature dimension computation arising in GNNs, in addition to vertex and edge parallelism.
% Prior graph processing systems employ sophisticated scheduling methods to tackle load imbalance caused by the irregularity in vertex or edge parallelism.
% We observe that while load imbalance tends to be the key factor prohibiting efficient parallel execution of tradition graph workloads on GPU, it is no longer a severe issue in GNNs with the new parallelization opportunities provided by the feature dimension.
% Instead, we argue that adapting parallelization strategies for computation patterns is the key to enabling performant processing of GNN workloads on GPU.

For vertex-wise computations, {\system} incorporates vertex parallelization into the SpMM template and allows users to specify a parallelization scheme for the message function with an FDS.
For example, the FDS for GCN aggregation is shown in Figure \ref{fig:spmm-code-gcn-conv} at line 19--22, which, combined with the SpMM template, defines the parallelization strategy shown in Figure \ref{fig:specialized-parallelization-gpu-sub1}: each CUDA block processes a number of vertices, which correspond to several rows in the adjacency matrix, and the feature dimension is parallelized across the threads in one CUDA block.
This simple parallelization strategy turns out to be highly efficient---there is no load imbalance within each CUDA block since all threads are assigned exactly the same amount of work; no control divergence; read requests into global memory from the threads within one CUDA block are contiguous and can be coalesced to realize high bandwidth utilization.
This parallelization strategy is first proposed in \cite{yang2018design} that focuses on manually optimizing the vanilla SpMM kernel; we can easily express it with the programming infrastructure of {\system} to optimize a broad class of generalized SpMM computations.

% One caveat in this method is that load imbalance may still occur across CUDA blocks due to the intrinsic irregularity in vertex parallelism, namely, vertices have differing numbers of neighbors.
% We argue that this cross-block load imbalance is not as severe an issue as cross-thread load imbalance that occurs in traditional graph workloads.
% Because CUDA blocks are executed independently of each other with the execution order determined by a hardware scheduler at run time, the compute capacity on GPUs can be fully utilized as long as enough CUDA blocks are launched.
% Empirical evaluation in Section \ref{sec:evaluation} confirms the effectiveness of this simple parallelization strategy without special handling of cross-block load imbalance.
% Though not critical, existing load balancing methods (e.g. splitting vertices with a large number of neighbors) are compatible with feature dimension parallelization and thus can be incorporated.

For edge-wise computations, {\system} incorporates edge parallelization into the SDDMM template and allows users to specify a parallelization scheme for the edge function with an FDS.
For example, the FDS for dot-product attention is shown in Figure \ref{fig:sddmm-code-dot-product-attention} at line 13--16, which, combined with the SDDMM template, defines the parallelization strategy shown in Figure \ref{fig:specialized-parallelization-gpu-sub2}: each CUDA block processes a number of edges, which correspond to several non-zero elements in the adjacency matrix, and all the threads in one CUDA block collectively process the dot-product operations on edges using tree reduction \cite{tree-reduce}.
Prior graph processing systems (e.g., Gunrock \cite{gunrock}), which are designed without being aware of feature dimension computation, fail to exploit this form of parallelism.

More complex UDFs that compute on multi-dimensional feature tensors require a multi-level parallelization scheme.
Figure \ref{fig:schedule-mlp-conv-gpu} shows the FDS for MLP aggregation---it parallelizes the first dimension across CUDA blocks and the second dimension across threads.

\subsubsection{Hybrid Partitioning on GPU}
The optimizations for graph traversal on CPU (e.g., 1D graph partitioning) are not directly applicable to GPU due to the differences between CPU and GPU memory architectures---shared memory size \revision{(configurable up to 96 KB on Tesla V100 GPU)} is much smaller than LLC, which is typically tens of Mega Bytes (MBs).
% These characteristics determine that on GPU a very large number of partitions are required, thus introducing significant overhead of merging intermediate results using atomic operations.
To make effective use of limited-capacity shared memory on GPU, we propose a hybrid partitioning method that processes high-degree vertices and low-degree vertices differently.
Specifically, this method reorders the vertices into a low-degree part and a high-degree part according to a threshold; it only partitions high-degree vertices and loads them to shared memory.
The intuition of hybrid partitioning is that high-degree vertices are accessed for more times and therefore can benefit more from shared memory optimization.
% This method is inspired by PowerLyra \cite{powerlyra}, a distributed graph processing system designed to handle the skewed distribution in natural graphs.
The key trade-off here is between read efficiency and merge cost---a smaller degree threshold leads to more partitions, which improves read efficiency but increases merge cost.

% \paragraph{Parameter Tuning.}
% In addition to specializing parallelization strategies for computation patterns and doing hybrid partitioning, achieving high-performance computing on GPUs requires carefully selecting parameters such as number of blocks and number of threads.
% These parameters incur intricate interplay between different types of hardware resource constraints (e.g. register usage, warp occupancy).
% In this work, parameter tuning is made easier by writing kernels as templates based on TVM compiler stack.
% We will elaborate on this in Section \ref{sec:implementation}.

% \input{figure_tex/fig_hybrid_partitioning_gpu.tex}

\section{System Implementation}
\label{sec:implementation}
% \yida{Implementation is two-fold: 1) integrating with DGL in practice; 2) utilizing TVM to implement the proposed kernel optimizations in \autoref{sec:design}. The description of TVM DSL and IR builder, and the discussion of potential auto-tuning in the future can go to this}
This section describes the implementation of {\system}, in particular, how we extended TVM to support the core sparse patterns of GNNs (i.e., generalized SpMM and SDDMM), and how we integrated {\system} into DGL.

\subsection{TVM IR Templates}
We implemented the SpMM and SDDMM templates as TVM IR templates.
TVM is a domain-specific language and compiler for tensor computations and has been widely adopted to accelerate deep learning workloads \cite{aws-cnn-cpu,aws-cnn-gpu}.
% TVM allows users to write tensor programs in a declarative paradigm, and provides a set of schedule primitives (e.g. loop split, loop reorder, vectorize) to apply optimizations.
Because TVM does not support sparse representation and computation in its tensor expression language, we implemented and optimized SpMM and SDDMM templates by directly constructing and manipulating the IR (intermediate representation) using lower-level APIs.
Feature dimension computations on each vertex/edge described by UDFs are dense and therefore easily supported.
{\system} combines scheduling parameters from the sparse templates (e.g., number of graph partitions, number of CUDA blocks) and those from the FDS (e.g., feature dimension tiling factors) to create the design space.
In this work we use na\"ive grid search to find the optimal parameters under a given input shape, and it is an interesting future direction to try more intelligent tuners \cite{opentuner, autotvm} for faster design space exploration.
After performing optimizations for both the templates and UDFs, {\system} inlines UDFs into the templates to generate fused kernels.

We parallelize the kernels over multiple threads on CPU using the customized thread pool \cite{aws-cnn-cpu} in TVM runtime, which is lightweight and particularly efficient in handling the kind of embarrassingly parallel workloads.
To avoid LLC contention after graph partitioning, we assign multiple threads to collectively work on one graph partition at a time instead of assigning each thread to a different partition.

\subsection{DGL Integration}
% DGL's scheduler automatically detects the pattern of fusion and dispatch them to specific executors.
% \system~supports all kind of operations that could be optimized by DGL.
% We replace the executors of SPMM and SDDMM in DGL with tunable TVM kernels provided by FeatGraph.
% \yida{The subsection is currently too short. We should elaborate how the integration works.}
In order to evaluate the performance of \system~in end-to-end GNN training and inference, we integrated \system~into DGL, a popular open-source GNN framework.
DGL implemented a minimal Gunrock-like graph kernel interface named Minigun \cite{minigun}.
With Minigun, DGL provided a set of builtin message functions and edge functions to support common GNN workloads.
For each of these builtin functions, we implemented a corresponding one with the programming infrastructure of {\system}, such as GCN aggregation and dot-product attention.
To handle more complex cases such as MLP aggregation, the current solution in DGL is to calculate and materialize the messages on every edge using deep learning systems as backend.
In contrast, {\system} generates fused kernels, thus both saving memory and improving efficiency.
{\system} generates kernel codes for a specific graph topology (i.e., the adjacency matrix); since GNN training typically involves hundreds of epochs, the compilation cost is amortized and negligible.

% DGL implemented a minimal Gunrock-like graph kernel interface that takes feature dimension into account without specifying optimization rules on these dimensions.
% DGL kernel optimizes GNN workloads of two kinds: 1. copy-reduce: copy node/edge features to edges and reduce on destination nodes; 2. binop(binary operation)-reduce: compute edge features by conduct binary operations on node/edge features then reduce on destination nodes. The binary operation phase correspond to edge-wise computations in \system while the reduce phase correspond to node-wise computations in \system.

% We replace the DGL kernels by rewriting copy-reduce and binop-reduce and their corresponding backward functions(used in back-propagation phase in GNN training) with vertex/edge-wise computations in \system. For each new graph structure, DGL dispatches the computation into specific copy-reduce and binop-reduce calls which have been transformed into \system functor calls. We build the forward/backward operator with TVM for each graph structure and cache this operator for future use. In generic training procedure of GNNs on a single GPU/CPU, we reuse the same graph but different inputs during the training phase, which means the kernel compile time is negligible for GNN training.

The integration requires a small amount of effort (only $\sim$ 300 lines of Python code) because both \system~and DGL follow the message passing paradigm in their programming interface design.
The integration with DGL demonstrates that it is straightforward to have \system~be the backend to accelerate GNN frameworks in general, including PyG, NeuGraph, etc.

\section{Evaluation}
\label{sec:evaluation}

This section seeks to answer the following questions:
\begin{enumerate}
    \itemsep0em
    \item What is the performance gain of GNN kernels on both CPU and GPU?
    \item What is the implication of each of our proposed optimization techniques for both templates and UDFs?
    \item \revision{Is the kernel performance sensitive to scheduling parameters and graph sparsity?}
    \item What is the speedup of end-to-end GNN training and inference brought by {\system} \revision{without affecting the accuracy of the models}?
\end{enumerate}

\subsection{Experiment Setup}

\textbf{Environment.}
For CPU evaluation, we conduct experiments on Amazon EC2 c5.9xlarge instance, which is a one-socket 18-core 3.0 GHz Intel Xeon Platinum 8124M machine with 25 MB LLC and 68 GB DRAM.
For GPU evaluation, we conduct experiments on p3.2xlarge instance, which has a Tesla V100 GPU with 80 SMs; \revision{each SM has shared memory configurable up to 96 KB (the default size is 48 KB)}.

\textbf{Datasets.}
% \yida{For ogbn-proteins, can you add one sentence of justification of why it is representative? Perhaps for reddit, too. Although we stated that it is commonly for evaluating accuracy of new models, we still don't explicitly say that why it is good for performance study.}
% \reply{I think by saying that reddit is commonly used for evaluating the accuracy of new models, we convey the information that this dataset is important and therefore improving the performance on this dataset is useful.}
Table \ref{tab:dataset} lists the datasets used for evaluation:
\revision{
\texttt{ogbn-proteins} represents proteins and their biological associations with vertices and edges---this dataset is from Open Graph Benchmark\footnote{\url{https://ogb.stanford.edu/}}, a realistic benchmark suite for GNNs;
\texttt{reddit} \cite{graphsage} is constructed from the Reddit online forum wherein vertices represent posts and edges are established if two posts are commented by a same use---this dataset is commonly used in GNN research for evaluating the accuracy of new models;
\texttt{rand-100K} is a synthetic graph wherein 20K vertices have an average degree of 2000 and the remaining 80K vertices have an average degree of 100---this dataset is specifically aimed at studying the effect of hybrid partitioning on GPU performance.
}

\textbf{Baselines.}
We compare \system~with state-of-the-art graph processing systems, specifically \emph{Ligra} on CPU and \emph{Gunrock} on GPU.
We also compare with vendor-provided sparse libraries, specifically \emph{MKL} (2019.5) on CPU and \emph{cuSPARSE} (10.1) on GPU whenever possible, as only a subset of GNN kernels are supported in these libraries.
\revision{In all the experiments, we first do a warm-up run and then take the average time of 10 runs as the measurement.}

\begin{table}[ht]
\centering
\begin{adjustbox}{width=0.8\linewidth}
\begin{tabular}{c|ccc}
\toprule
Graph dataset            & $|\mathcal{V}|$ & $|\mathcal{E}|$ & Average degree \\
\midrule
\texttt{ogbn-proteins}            & 132.5K  & 79.1M   & 597  \\
\texttt{reddit}                   & 233.0K  & 114.8M  & 493      \\
\texttt{rand-100K}                & 100.0K  & 48.0M   & 480  \\
\bottomrule
\end{tabular}
\end{adjustbox}
\caption{Graph datasets (K: thousand, M: million).}
\label{tab:dataset}
\vspace{0.5em}
\end{table}

\subsection{Performance Gain of GNN Kernels}

\begin{table}[t]

\begin{subtable}[ht]{\linewidth}
\centering
\begin{adjustbox}{width=1.0\linewidth}
\begin{tabular}{c|c|ccccc}
\toprule
\multicolumn{2}{c|}{\multirow{2}{*}{Unit: sec}} & \multicolumn{5}{c}{Feature length} \\
\multicolumn{2}{c|}{} & 32 & 64 & 128 & 256 & 512 \\
\midrule
\multicolumn{1}{c|}{\multirow{3}{*}{\texttt{ogbn-proteins}}}
& Ligra    	&	1.47	&	2.05	&	3.10	&	6.01	&	12.30	\\
& MKL    	&	0.60	&	\textbf{0.96}	&	2.17	&	5.34	&	14.71	\\
& FeatGraph	&	\textbf{0.50}	& 0.99	&	\textbf{1.97}	&	\textbf{3.94}	&	\textbf{8.02} \\
\midrule
\multicolumn{1}{c|}{\multirow{3}{*}{\texttt{reddit}}}
& Ligra	    &	4.10	&	7.20	&	13.10	&	20.40	&	34.90	\\
& MKL     	&	1.50	&	3.01	&	7.87	&	17.79	&	40.06	\\
& FeatGraph	&\textbf{1.02}	&\textbf{2.13}	&\textbf{4.09}	&\textbf{8.16}	&\textbf{16.71}	\\
\midrule
\multicolumn{1}{c|}{\multirow{3}{*}{\texttt{rand-100K}}}
& Ligra      & 0.64	   &    0.86	&   1.49	&   2.58	&   4.91 \\
& MKL        & 0.43    &	0.77	&   2.26	&   5.45	&   15.51 \\
& FeatGraph  & \textbf{0.22}  & \textbf{0.43}	&  \textbf{0.87} & \textbf{1.74} & \textbf{3.52} \\
\bottomrule
\end{tabular}
\end{adjustbox}
\caption{GCN aggregation}
\end{subtable}

\vspace{.2cm}

\begin{subtable}[ht]{\linewidth}
\centering
\begin{adjustbox}{width=1.0\linewidth}
\begin{tabular}{c|c|ccccc}
\toprule
\multicolumn{2}{c|}{\multirow{2}[0]{*}{Unit: sec}} & \multicolumn{5}{c}{Feature length} \\
\multicolumn{2}{c|}{} & 32 & 64 & 128 & 256 & 512 \\
\midrule
\multicolumn{1}{c|}{\multirow{2}{*}{\texttt{ogbn-proteins}}}
& Ligra   	 & 12.90 	    & 24.70        & 47.70        & 94.00         &187.00	\\
& FeatGraph  &\textbf{2.48} &\textbf{4.84} &\textbf{9.68} &\textbf{19.55} &\textbf{38.70}\\
\midrule
\multicolumn{1}{c|}{\multirow{2}{*}{\texttt{reddit}}}
& Ligra 	 & 20.70	    & 37.90         & 71.50         & 139.00        & 273.00 \\
& FeatGraph	 &\textbf{4.03} &\textbf{8.20}	&\textbf{15.33}	&\textbf{30.80}	&\textbf{62.07}	\\
\midrule
\multicolumn{1}{c|}{\multirow{2}{*}{\texttt{rand-100K}}}
& Ligra      & 7.81 	    & 14.80        & 28.80        & 56.90         & 113.00	\\
& FeatGraph  &\textbf{1.42} &\textbf{2.74} &\textbf{5.48} &\textbf{10.96} &\textbf{21.97}\\
\bottomrule
\end{tabular}
\end{adjustbox}
\caption{MLP aggregation}
\end{subtable}

\vspace{.2cm}

\begin{subtable}[ht]{\linewidth}
\centering
\begin{adjustbox}{width=1.0\linewidth}
\begin{tabular}{c|c|ccccc}
\toprule
\multicolumn{2}{c|}{\multirow{2}[0]{*}{Unit: sec}} & \multicolumn{5}{c}{Feature length} \\
\multicolumn{2}{c|}{} & 32 & 64 & 128 & 256 & 512 \\
\midrule
\multicolumn{1}{c|}{\multirow{2}{*}{\texttt{ogbn-proteins}}}
& Ligra	    &	9.81	&	22.30	&	47.50	&	97.70	&	198.00	\\
& FeatGraph	& \textbf{2.21}	&\textbf{4.39}	&\textbf{8.67}	&\textbf{16.46}	&\textbf{32.97}	\\
\midrule
\multicolumn{1}{c|}{\multirow{2}{*}{\texttt{reddit}}}
& Ligra	    &	17.20	&	37.30	&	77.20	&	152.00	&	297.00	\\
& FeatGraph &\textbf{3.71}	&\textbf{7.34}	&\textbf{14.11}	&\textbf{27.13}	&\textbf{54.51}	\\
\midrule
\multicolumn{1}{c|}{\multirow{2}{*}{\texttt{rand-100K}}}
& Ligra      &  5.57	&   12.90	&   28.20	&   58.30	&   119.00 \\
& FeatGraph  &  \textbf{1.28}	&\textbf{2.51}	& \textbf{5.37}	& \textbf{10.76} &\textbf{21.47} \\
\bottomrule
\end{tabular}
\end{adjustbox}
\caption{Dot-product attention}
\end{subtable}

\caption{Single-threaded CPU performance. Best result is marked in \textbf{bold}.}
\label{tab:cpu-kernel}
\vspace{-1em}
\end{table}

\begin{figure}[t]
\centering

\includegraphics[width=0.7\linewidth]{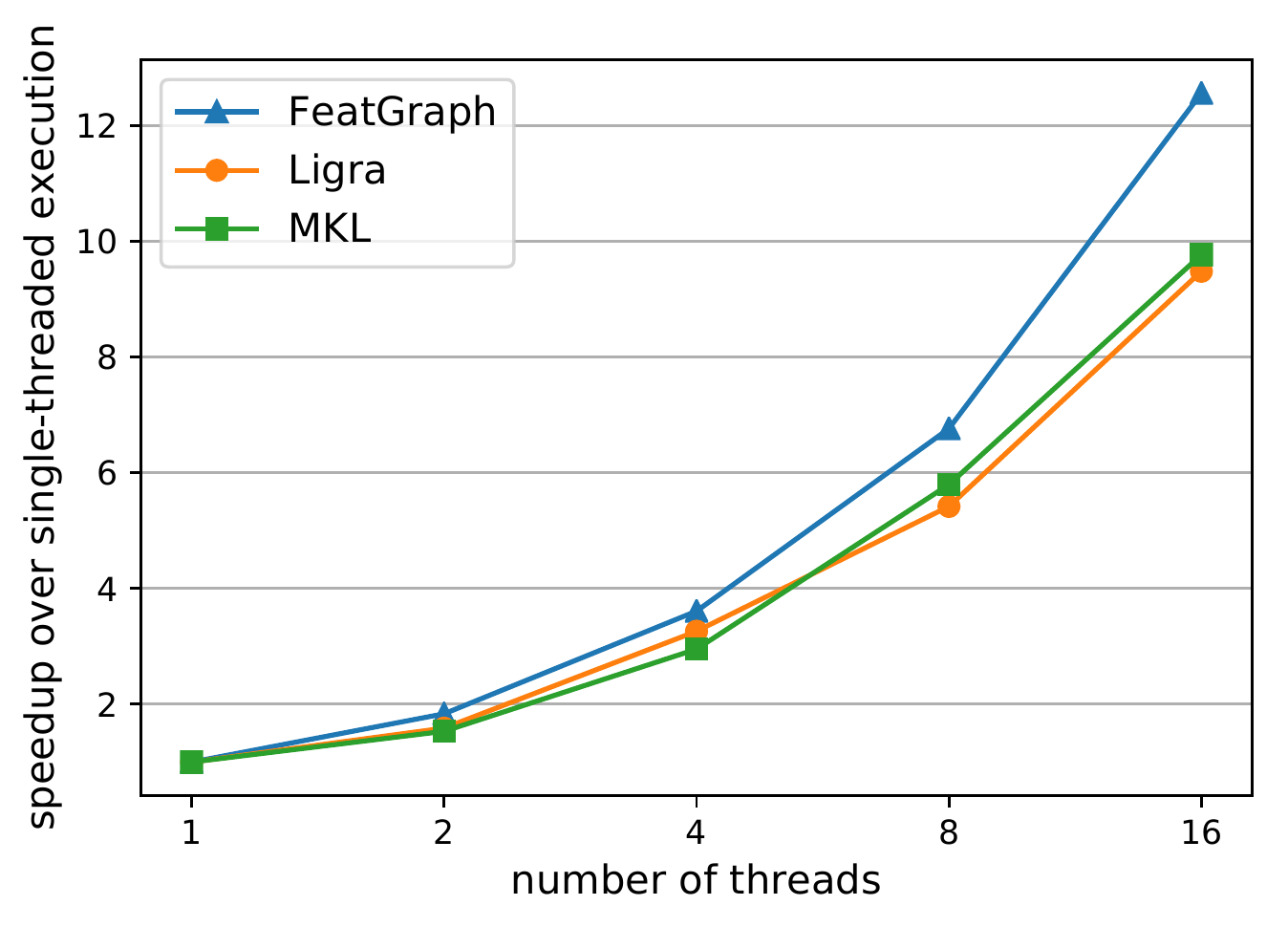}

\caption{Scalability comparison of FeatGraph with Ligra and MKL. Evaluated on GCN aggregation. Tested on \texttt{reddit} with feature length 512.}
\label{fig:result-cpu-kernel-SpMM-multi-threads}
% \vspace{0.5em}
\end{figure}

\begin{table}[t]

\begin{subtable}[h]{1.0\linewidth}
\centering
\begin{adjustbox}{width=1.\linewidth}
\begin{tabular}{c|c|ccccc}
\toprule
\multicolumn{2}{c|}{\multirow{2}{*}{Unit: ms}} & \multicolumn{5}{c}{Feature length} \\
\multicolumn{2}{c|}{} & 32 & 64 & 128 & 256 & 512 \\
\midrule
\multicolumn{1}{c|}{\multirow{3}{*}{\texttt{ogbn-proteins}}}
& Gunrock	&  114.2  &	276.7	 & 1322.3	& 4640.3	& 12423.9	\\
& cuSPARSE	& \textbf{4.1} & 8.1 	 & 16.2 	& 32.1	    & 64.2      \\
& FeatGraph	& 4.6 & \textbf{7.8}  & \textbf{15.4}	& \textbf{30.8}	  & \textbf{61.9}  \\
\midrule
\multicolumn{1}{c|}{\multirow{3}{*}{\texttt{reddit}}}
& Gunrock   &  616.9  & 2026.4	 & 5141.2	& 11715.3	& 24749.8	\\
& cuSPARSE	&  \textbf{12.2}  &	\textbf{25.1} & \textbf{51.6}& \textbf{104.7} & \textbf{209.6}	\\
& FeatGraph	&  14.3   &	28.6	 & 57.8 	& 116.9 	& 232.0	\\
\midrule
\multicolumn{1}{c|}{\multirow{3}{*}{\texttt{rand-100K}}}
& Gunrock   &  72.7	  & 175.5 	 & 1006.2 	& 3303.7	& 8236.5 \\
& cuSPARSE  &  3.6	  & 5.9	     & 10.6	    & 21.9	    & 44.4 \\
& FeatGraph &  \textbf{2.8}    &\textbf{4.9}    & \textbf{10.2}	  & \textbf{20.3}& \textbf{39.9} \\
\bottomrule
\end{tabular}
\end{adjustbox}
\caption{GCN aggregation}
\end{subtable}

\vspace{.2cm}

\begin{subtable}[h]{\linewidth}
\centering
\begin{adjustbox}{width=1.0\linewidth}
\begin{tabular}{c|c|ccccc}
\toprule
\multicolumn{2}{c|}{\multirow{2}{*}{Unit: ms}} & \multicolumn{5}{c}{Feature length} \\
\multicolumn{2}{c|}{} & 32 & 64 & 128 & 256 & 512 \\
\midrule
\multicolumn{1}{c|}{\multirow{2}{*}{\texttt{ogbn-proteins}}}
& Gunrock	 & 591.6        & 833.4	       & 2067.7       & 5603.5        & 13687.4	\\
& FeatGraph  &\textbf{26.9} &\textbf{46.7} &\textbf{87.4} &\textbf{168.9} &\textbf{332.9}\\
\midrule
\multicolumn{1}{c|}{\multirow{2}{*}{\texttt{reddit}}}
& Gunrock   & 1285.6       & 2697.5	      & 5886.4	      & 12285.0	      & 25442.3	\\
& FeatGraph &\textbf{33.2} &\textbf{76.7} &\textbf{142.9} &\textbf{277.1} &\textbf{547.9}\\
\midrule
\multicolumn{1}{c|}{\multirow{2}{*}{\texttt{rand-100K}}}
& Gunrock    & 447.2       & 648.1        & 1556.1       & 3848.5       & 8624.6\\
& FeatGraph  &\textbf{8.9} &\textbf{14.9} &\textbf{26.0} &\textbf{46.6} &\textbf{89.6}\\
\bottomrule
\end{tabular}
\end{adjustbox}
\caption{MLP aggregation}
\end{subtable}

\vspace{.2cm}

\begin{subtable}[h]{\linewidth}
\centering
\begin{adjustbox}{width=1.0\linewidth}
\begin{tabular}{c|c|ccccc}
\toprule
\multicolumn{2}{c|}{\multirow{2}{*}{Unit: ms}} & \multicolumn{5}{c}{Feature length} \\
\multicolumn{2}{c|}{} & 32 & 64 & 128 & 256 & 512 \\
\midrule
\multicolumn{1}{c|}{\multirow{2}{*}{\texttt{ogbn-proteins}}}
& Gunrock	& 30.9	  & 58.8	& 120.2 	& 251.3 	& 645.1	\\
& FeatGraph	& \textbf{24.4}	 & \textbf{37.9} & \textbf{69.3} & \textbf{143.3} & \textbf{333.7} \\
\midrule
\multicolumn{1}{c|}{\multirow{2}{*}{\texttt{reddit}}}
& Gunrock	& 44.8	  & 99.3	& 278.5 	& 648.2 	& 1388.7	\\
& FeatGraph	& \textbf{35.9} & \textbf{56.6}	& \textbf{103.7}& \textbf{212.0} & \textbf{483.2} 	\\
\midrule
\multicolumn{1}{c|}{\multirow{2}{*}{\texttt{rand-100K}}}
& Gunrock   & 19.3	  & 37.3	& 75.5  	& 174.3	    & 441.6 \\
& FeatGraph & \textbf{14.9}	& \textbf{23.2}	& \textbf{42.3}	  & \textbf{87.8} & \textbf{201.5} \\
\bottomrule
\end{tabular}
\end{adjustbox}
\caption{Dot-product attention}
\end{subtable}

\caption{GPU performance. Best result is marked in \textbf{bold}.}
\label{tab:gpu-kernel}
\vspace{-0.5em}
\end{table}

\begin{figure*}[t]

\minipage{0.325\textwidth}
\includegraphics[width=1.0\linewidth]{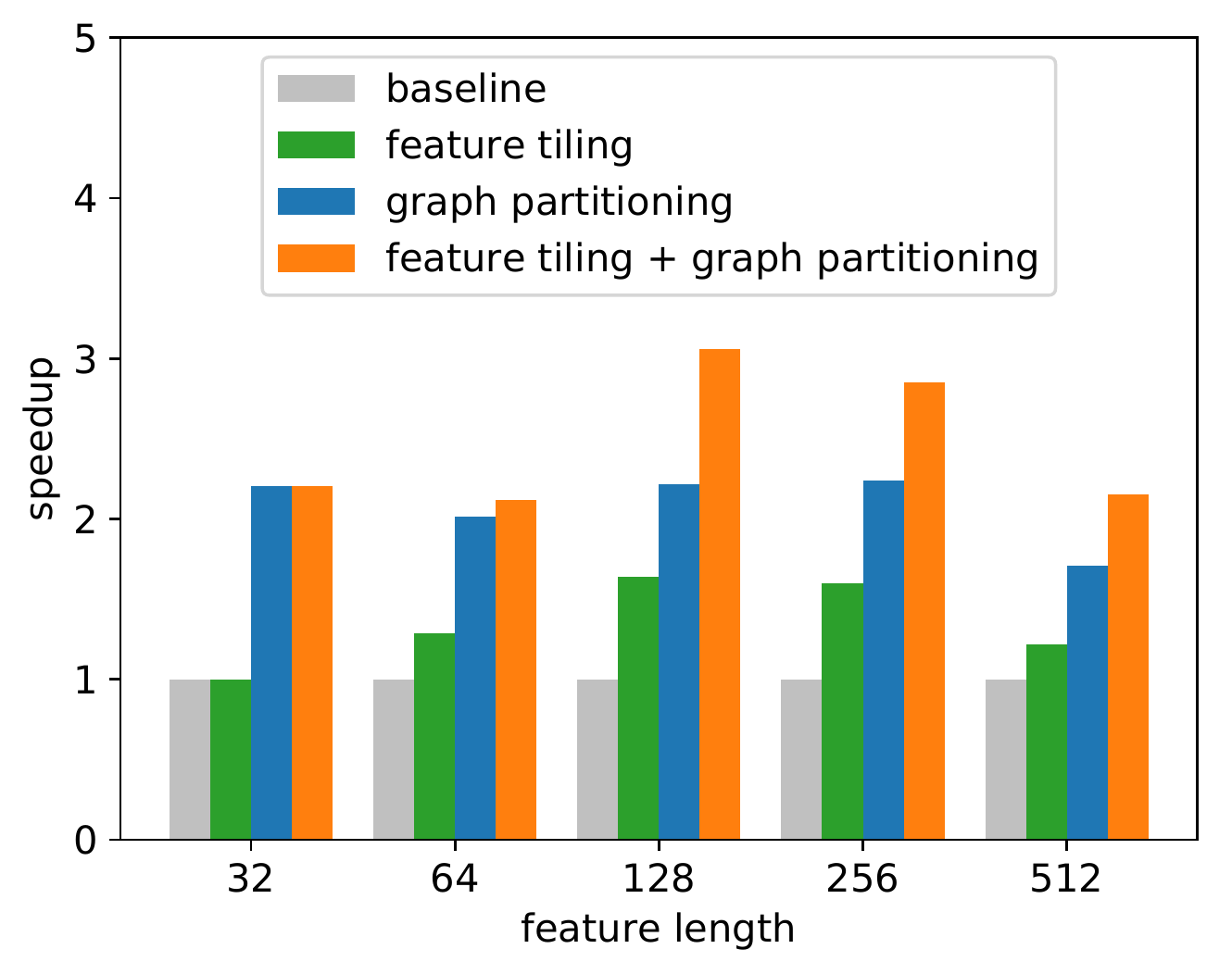}
\caption{Effect of graph partitioning and feature tiling on the CPU performance of GCN aggregation. Tested on \texttt{reddit}.}
\label{fig:effect-tiling-cpu-spmm}
\endminipage\hfill
\minipage{0.335\textwidth}
\includegraphics[width=1.0\linewidth]{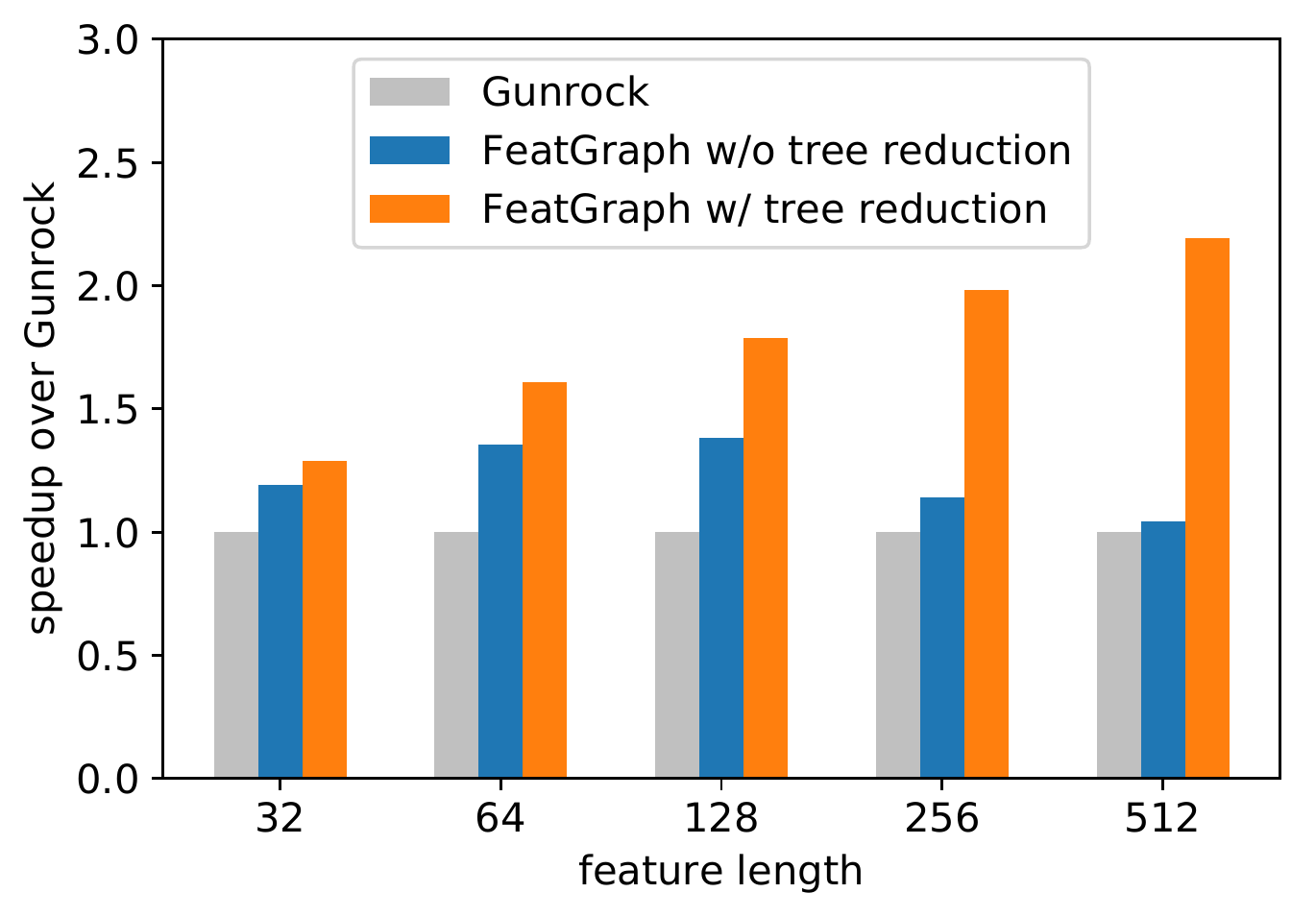}
\caption{Effect of tree reduction on the GPU performance of dot-product attention. Tested on \texttt{rand-100K}.}
\label{fig:effect-tree-reduction-gpu-sddmm}
\endminipage\hfill
\minipage{0.31\textwidth}
\includegraphics[width=1.0\linewidth]{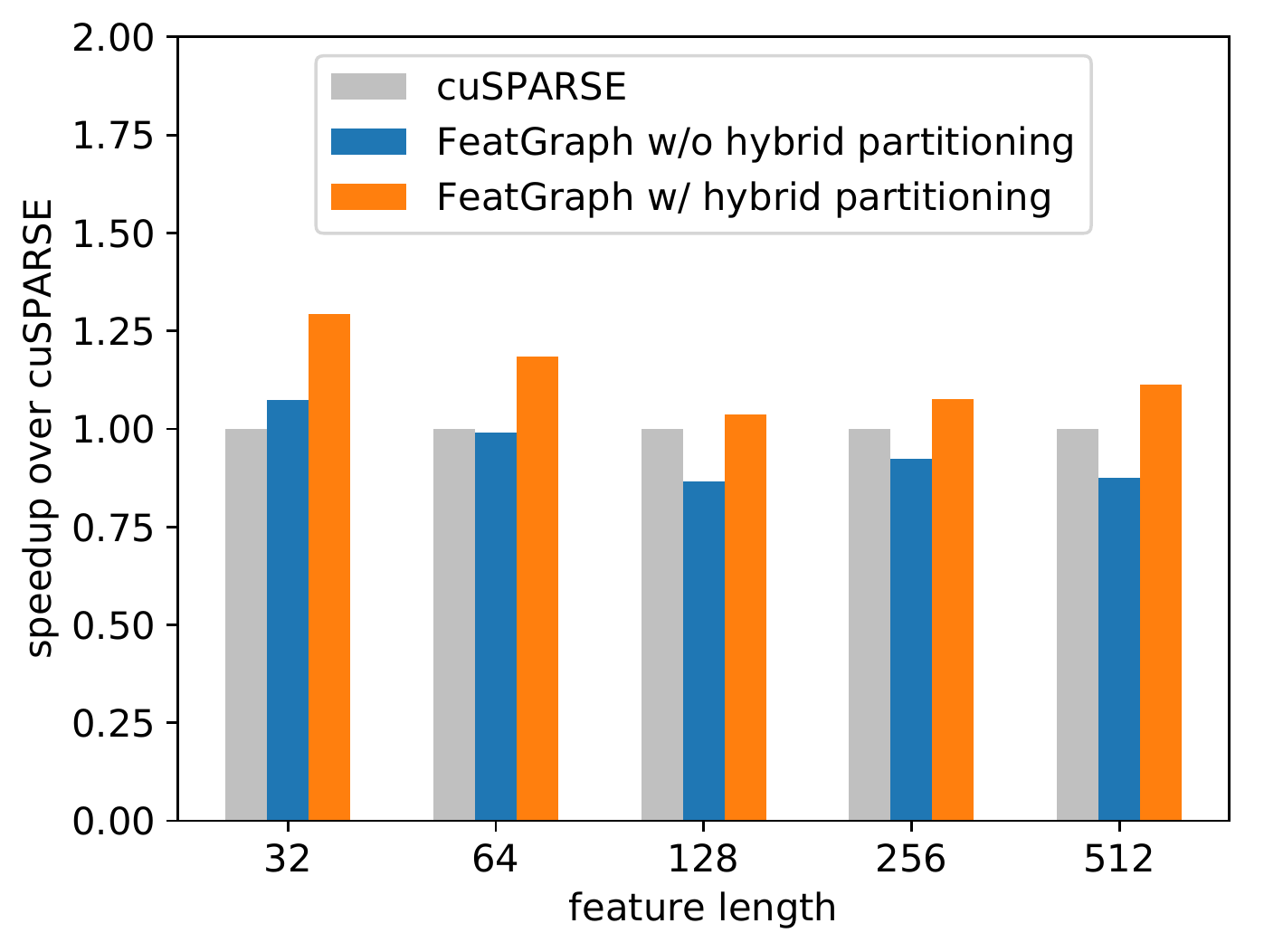}
\caption{Effect of hybrid partitioning on the GPU performance of GCN aggregation. Tested on \texttt{rand-100K}.}
\label{fig:effect-hybrid-partitioning-gpu-spmm}
\endminipage
\vspace{-1em}
\end{figure*}

We evaluate the performance gain of \system~on three kernels: GCN aggregation, MLP aggregation, and dot-product attention.
\revision{The kernels are performed on the full graph.}
We do the evaluation across a spectrum of feature lengths.
For MLP aggregation, the feature length refers to \texttt{d2} that is shown in Figure \ref{fig:spmm-code-mlp-conv}; \texttt{d1} is fixed as 8.

\textbf{Single-threaded CPU Performance.}
Table \ref{tab:cpu-kernel} shows that across all the evaluated datasets under different feature lengths, \system~achieves 1.4$\times$--4.0$\times$ speedup over Ligra on GCN aggregation, 4.4$\times$--5.5$\times$ speedup on MLP aggregation, and 4.3$\times$--6.0$\times$ speedup on dot-product attention, using a single thread.
Compared against MKL on GCN aggregation, \system~is faster in 14 out of 15 cases and achieves higher speedup with a larger feature length.
Specifically, when the feature length is 512, \system~is 1.8$\times$ faster on \texttt{ogbn-proteins}, 2.4$\times$ faster on \texttt{reddit}, and 4.4$\times$ faster on \texttt{rand-100K}.
MKL does not support MLP aggregation and dot-product attention.

% Mention that MKL's inspector-executor optimization is only for block-structured adjacency matrix, not useful here

\textbf{Multi-threaded CPU Performance.}
Figure \ref{fig:result-cpu-kernel-SpMM-multi-threads} shows that with 16 threads, for GCN aggregation on \texttt{reddit}, \system~achieves 12.6$\times$ speedup over its single-threaded execution, which is slightly higher than Ligra (9.5$\times$) and MKL (9.8$\times$).
Similar observation applies to other datasets and kernels.
As a result, \system~outperforms the others consistently in multi-threaded environment.
\system~scales well due to two factors: 1) its parallelization method avoids LLC contention by assigning multiple threads to collectively work on one graph partition at a time; 2) the thread pool in TVM runtime is lightweight \cite{aws-cnn-cpu}.

\textbf{GPU Performance.}
Table \ref{tab:gpu-kernel} shows that \system~is 24$\times$--206$\times$ faster than Gunrock on GCN aggregation, 18$\times$--96$\times$ faster on MLP aggregation, and 1.2$\times$--3.1$\times$ faster on dot-product attention.
The extreme slowness of Gunrock on GCN aggregation and MLP aggregation is caused by two reasons: 1) Gunrock's edge parallelization execution incurs huge overhead of atomic operations for vertex-wise reductions such as GCN aggregation and MLP aggregation; 2) Gunrock fails to exploit parallelism in feature dimension computation.
\system~is on par with cuSPARSE on GCN aggregation, being 10\%--20\% faster on \texttt{ogbn-proteins} and \texttt{rand-100K} while 10\% slower on \texttt{reddit}.
Notably, cuSPARSE does not support MLP aggregation and dot-product attention.\footnote{\final{The latest cuSPARSE supports dot-product attention via ConstrainedGeMM.}}

\subsection{Optimization Implications}

This subsection investigates the performance boost of each individual optimization technique described in Section \ref{sec:design}.
For the sake of space, in each ablation analysis we only pick one dataset to show the optimization effects.
Other datasets share similar observations.

\textbf{Graph Partitioning and Feature Dimension Tiling.}
Figure \ref{fig:effect-tiling-cpu-spmm} shows that feature dimension tiling combined with graph partitioning effectively boosts the performance of GCN aggregation on CPU.
Specifically, when the feature length is 512, feature dimension tiling alone and graph partitioning alone bring 1.2$\times$ speedup and 1.7$\times$ speedup, respectively; combining two achieves 2.2$\times$ speedup.

\textbf{Adaptive Parallelization Strategies on GPU.}
Figure \ref{fig:effect-tree-reduction-gpu-sddmm} shows that tree reduction boosts the performance of dot-product attention by up to 2$\times$.
The na\"ive parallelization strategy in Gunrock that assigns the entire dot product operation on each edge to one CUDA thread is less efficient in handling large feature lengths due to consuming too many registers per thread.

\textbf{Hybrid Partitioning on GPU.}
Figure \ref{fig:effect-hybrid-partitioning-gpu-spmm} shows the effect of hybrid partitioning on GCN aggregation tested on \texttt{rand-100}.
\system~gets 10\%--20\% performance boost by hybrid partitioning, and consequently outperforms cuSPARSE.

\vspace{-1em}
\revision{\subsection{Sensitivity Analysis}}

\begin{figure}[t]
\centering

\includegraphics[width=0.75\linewidth]{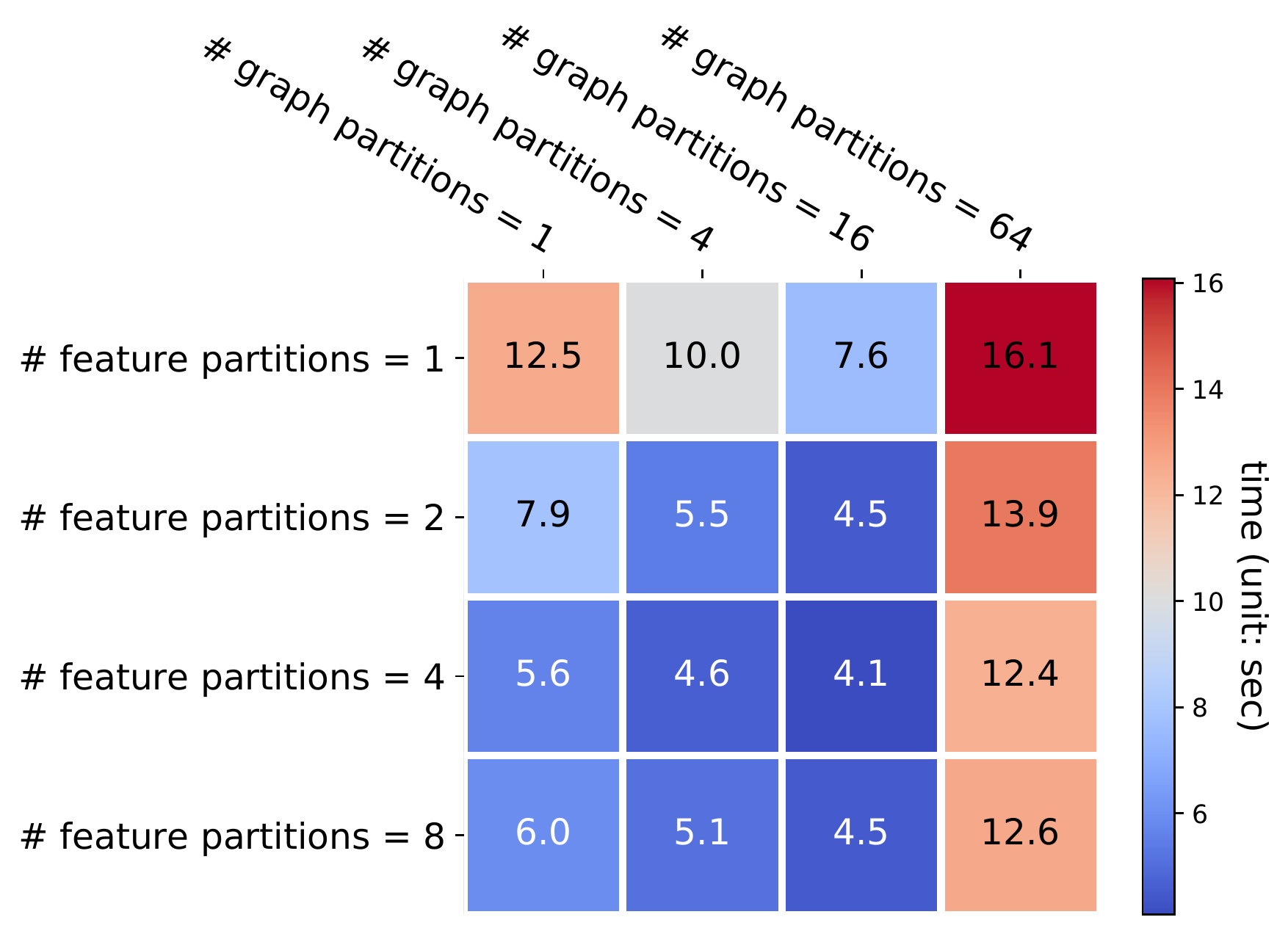}

\caption{Sensitivity of {\system} performance to partitioning factors for GCN aggregation on CPU. The dataset is \texttt{reddit}. The feature length is 128.}
\label{fig:sensitivity-partitioning-factors}
% \vspace{-1em}
\end{figure}

\revision{\textbf{Sensitivity to Partitioning Factors.}
Figure \ref{fig:sensitivity-partitioning-factors} shows that the performance of {\system} is sensitive to partitioning factors for GCN aggregation on CPU.
Specifically, on \texttt{reddit}, when the feature length is 128, the best performance is achieved with 16 graph partitions and 4 feature partitions.
On the same graph, as the feature length increases, the optimal number of feature partitions increases proportionately, while the optimal number of graph partitions stays constant.
% Other kernels of other datasets on both CPU and GPU share similar observation.
% \reply{Partitioning is only useful on CPU; On GPU, it's mainly about parallelization, which will be discussed in the next paragraph. I don't think we have to explicitly claim that other kernels of other datasets share similar observation.}
Transferable tuning across graphs, i.e., using the optimal partitioning factors tuned on one graph to predict the optimal partitioning factors for a new graph, is more challenging and worth further study.
}

\begin{figure}[t]
\centering

\includegraphics[width=0.75\linewidth]{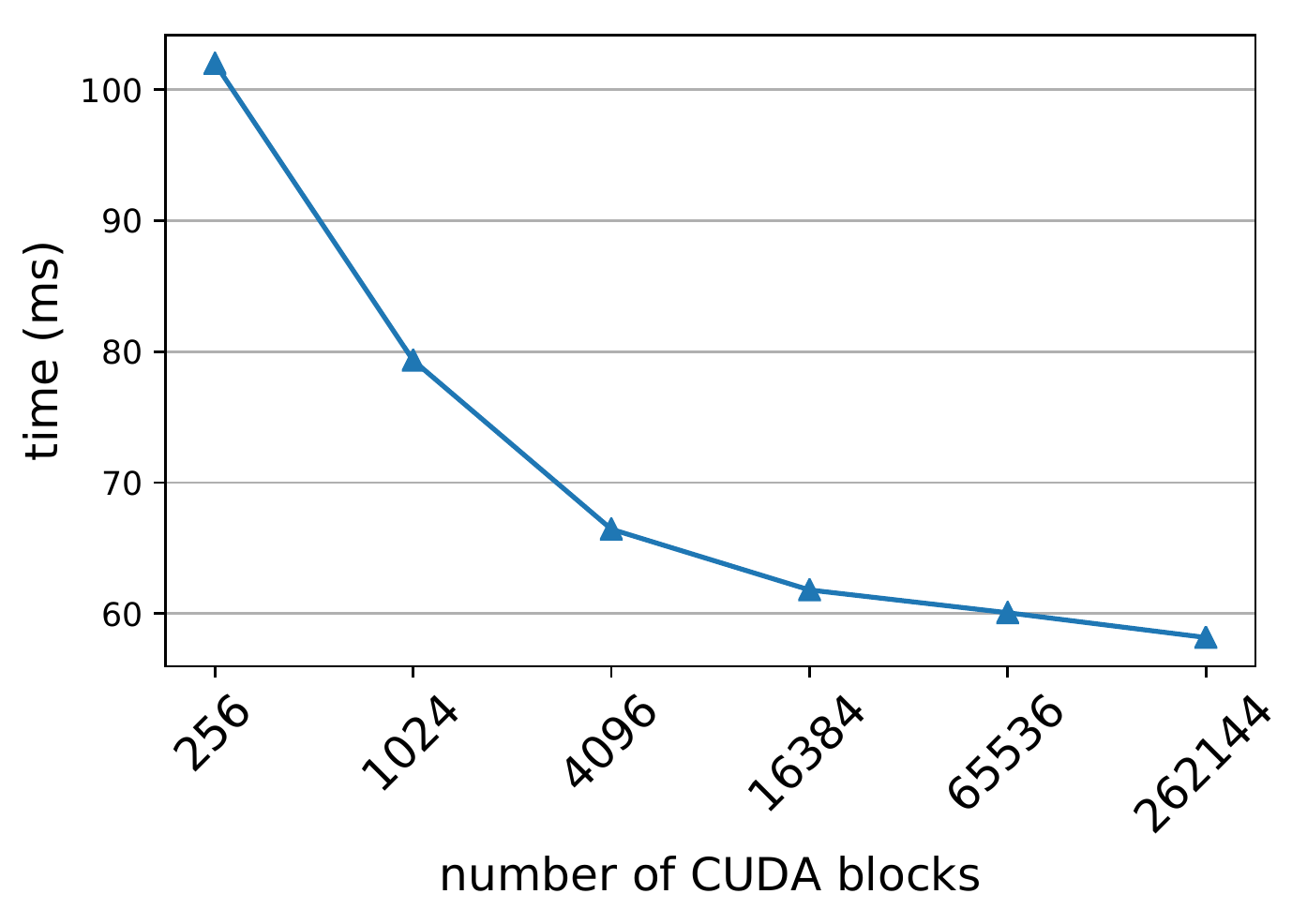}

\caption{Sensitivity of {\system} performance to the number of CUDA blocks for GCN aggregation on GPU. The dataset is \texttt{reddit}. The feature length is 128.}
\label{fig:sensitivity-spmm-number-cuda-blocks}
\vspace{-0.5em}
\end{figure}

\revision{
\textbf{Sensitivity to GPU Parameters.}
Figure \ref{fig:sensitivity-spmm-number-cuda-blocks} shows that {\system} performs better with a larger number of CUDA blocks for GCN aggregation on GPU, because a larger number of CUDA blocks can better utilize the massive parallel compute capacity of GPU.
In the evaluation, we set the number of CUDA blocks to the number of rows of the adjacency matrix.
}

\revision{
\textbf{Sensitivity to Graph Sparsity.}
Table \ref{tab:sensitivity-graph-sparsity} shows that {\system} achieves higher speedup over MKL as the graph sparsity decreases for GCN aggregation on CPU.
This trend is because a denser graph has more data reuse, which {\system} is able to exploit by graph partitioning and feature dimension tiling.
}

\begin{table}[ht]
\centering
\begin{adjustbox}{width=0.95\linewidth}
\begin{tabular}{c|c|c|c}
\toprule
Graph sparsity    & MKL (unit: sec) & {\system} (unit: sec) & Speedup \\
\midrule
99.95\%            & 0.34            & 0.31                  & 1.10$\times$  \\
99.5\%             & 3.58            & 1.95                  & 1.84$\times$      \\
95\%               & 37.22           & 12.78                 & 2.91$\times$  \\
\bottomrule
\end{tabular}
\end{adjustbox}
\caption{Sensitivity of {\system} performance to graph sparsity for GCN aggregation on CPU. The dataset is a synthetic uniform graph with 100K vertices. The feature length is 128.}
\label{tab:sensitivity-graph-sparsity}
\end{table}

% \yida{We should answer reviewer 1's question about ``How much tuning and analysis must be done for parameter tuning?'' somewhere in the paper.}
% \reply{addressed in the last paragraph of Section V-E}

% \yida{Given the extra page and the reviewer's interest in parameter tuning, we can consider bringing the paragraph of ``Sensitivity to GPU Parameters'' back.}
% \reply{Good suggestion. Added}

\vspace{1em}
\subsection{End-To-End GNN Training and Inference}

\begin{table}[tb]
\centering
\begin{adjustbox}{width=0.9\linewidth}
\begin{tabular}{c|c|c|c|c}
\toprule

\multicolumn{2}{c|}{\multirow{3}{*}{}} & \multirow{3}{*}{\shortstack{DGL w/o\\{\system}\\ (unit: sec)}} & \multirow{3}{*}{\shortstack{DGL w/\\{\system}\\(unit: sec)}} & \multirow{3}{*}{Speedup} \\
\multicolumn{2}{c|}{} & & & \\
\multicolumn{2}{c|}{} & & & \\
\midrule
\multicolumn{1}{c|}{\multirow{3}{*}{\shortstack{CPU\\training}}}
& GCN        	&  2447.1	&	114.5   & 21.4$\times$	\\
& GraphSage   	&  1269.6	&	57.8    & 21.9$\times$	\\
& GAT	        &  5763.9 	&   179.3	& 32.2$\times$	 \\
\midrule
\multicolumn{1}{c|}{\multirow{3}{*}{\shortstack{CPU\\inference}}}
& GCN   	    &  1176.9	&  55.3    &	21.3$\times$   \\
& GraphSage    	&   602.4	&  29.8    &	20.2$\times$	\\
& GAT			&  1580.9  	&   71.5   &    22.1$\times$  \\
\midrule
\multicolumn{1}{c|}{\multirow{3}{*}{\shortstack{GPU\\training}}}
& GCN    	    &	6.3  	&	2.2     &	  2.9$\times$ \\
& GraphSage    	&	3.1	    &   1.5     &     2.1$\times$	\\
& GAT			&   *N/A    &   1.64    &     *N/A \\
\midrule
\multicolumn{1}{c|}{\multirow{3}{*}{\shortstack{GPU\\inference}}}
& GCN    	    &	3.1  	&	1.5  &	   2.1$\times$ \\
& GraphSage    	&	1.5	    &	1.1  &	   1.4$\times$\\
& GAT			&   8.1	    &	1.1  &     7.1$\times$\\
\bottomrule
\end{tabular}
\end{adjustbox}

\caption{Speedup of end-to-end GNN training and inference brought by {\system}. Tested on \texttt{reddit}. Time is for one epoch. (*GAT training in DGL w/o \system runs out of GPU memory.)}
\vspace{-1em}
\label{tab:end2end}
\end{table}

We integrated \system~into DGL and evaluated the performance of \system~in end-to-end GNN training and inference on three models: a 2-layer graph convolutional network (GCN) \cite{gcn} of hidden size 512, a 2-layer GraphSage \cite{graphsage} of hidden size 256, and a 2-layer graph attention network (GAT) \cite{gat} of hidden size 256.
GCN uses sum aggregation and requires generalized SpMM computations in both forward and backward propagation; GraphSage follows a similar architecture as GCN but allows more flexible aggregation functions (e.g., max); GAT uses dot-product attention, thus requiring both generalized SpMM and SDDMM computations.

\textbf{Accuracy.}
\revision{
{\system} as a backend is for performance optimization without changing the semantics of GNN models.
As a sanity check, we evaluate the accuracy of the three models on the task of vertex classification.
The 233K vertices of the \texttt{reddit} dataset are split into 153K, 24K, and 56K for training, validation, and testing, respectively.
We train the models for 200 epochs.
The testing accuracy obtained by DGL using {\system} matches that obtained by DGL using its original backend Minigun---93.7\% for GCN and 93.1\% for GraphSage.
The training of GAT does not converge due to gradient explosion, with either {\system} backend or Minigun backend.
}

\textbf{Speedup.}
\revision{
Table~\ref{tab:end2end} reports the training and inference time of one epoch for the three GNN models.
The time of tuning partitioning factors is excluded, because it is amortized over multiple epochs---it is less than 1\% of the time of training GCN for 200 epochs.
Furthermore, the partitioning factors tuned on GCN are directly applied to GraphSage and GAT---the number of graph partitions is kept the same and the number of feature partitions is adjusted to the feature length.}
The results show that on CPU, \system~speeds up both training and inference by more than 20$\times$ on all the three models; on GPU, \system~speeds up training by more than 2$\times$, and inference by 1.4$\times$--7.1$\times$.
The highest speedup is achieved on GAT, which has a more complex architecture than GCN and GraphSage.

\section{Related Work}
\label{sec:related}

Recent years have seen an emergence of specialized frameworks that attempt to make the processing of GNN workloads easier and faster.
For example, DGL \cite{dgl} and PyG \cite{PyG} wrap deep learning systems with a message-passing programming interface for GNNs.
NeuGraph \cite{neugraph} addresses the challenge of large-scale GNN training by partitioning the dataflow over multiple GPUs and employing a chain-based streaming schedule.
\system~focuses on optimizing graph-specific kernels, and can be integrated into these GNN frameworks to serve as an efficient backend on both CPU and GPU.

Systems for processing traditional graph workloads (e.g., BFS, PageRank) have been extensively studied in literature~\cite{pregel, x-stream, powergraph, ligra, gunrock, flashgraph}.
These systems allow users to flexibly express graph algorithms by defining computation on each vertex/edge.
Among them, Ligra \cite{ligra} achieves superior performance on CPU by dynamically switching the message propagation direction (i.e., push or pull) based on the size of the frontier (active vertices) at each iteration, and Gunrock \cite{gunrock} achieves superior GPU performance by sophisticated scheduling methods to ensure load balance in its edge parallelization execution.
However, Ligra is not exploiting cache optimization, and its push-pull optimization is no longer critical in GNN workloads since typically all vertices are active at each layer of a GNN model.
Gunrock fails to achieve good performance for GNN workloads because it is unable to exploit parallelism in feature dimension computation, let alone adapt parallelization strategies for computation patterns.

There is another series of works that focus on formulating graph algorithms as sparse linear algebra operations \cite{graphblas, graphmat, graphlinearalgebra, semispmm}.
For example, BFS is formulated as a sequence of sparse matrix sparse vector multiplication (SpMSpV); PageRank is formulated as a sequence of sparse matrix dense vector multiplication (SpMV).
\system~borrows from these works the general idea of mapping graph computations to sparse kernels.
\system~differs from these works in two major aspects: 1) \system~can express more complex user-defined functions (UDFs) to support the diverse variants of GNN models; 2) \system~pays a special attention to optimizations of feature dimension computation, which are unexploited in previous efforts.

Vendor-specific libraries (e.g., MKL \cite{mkl} and cuSPARSE \cite{cusparse}) provide highly optimized implementations for sparse kernels that are identified important to a broad range of applications.
Compared with these libraries, {\system} is more comprehensive at kernel coverage for GNN's use case.
Besides, by adopting a tensor compiler approach in contrast to the manual optimization approach of these libraries, {\system} is able to search for the best scheduling schemes on both CPU and GPU.

% A variety of graph partitioning methods have been proposed to improve cache utilization \cite{cagra}, to reduce memory-disk data transfer in out-of-core systems \cite{gridgraph}, or to reduce communication in distributed settings \cite{3D-partitioning}.
% \system is the first to co-optimize graph traversal and vertex-wise/edge-wise computations for cache optimization.

TACO \cite{taco} is a compiler targeting general sparse tensor computations by the design of a flexible sparse tensor representation system.
However, TACO does not allow scheduling as TVM, and it lacks support for generating high-quality GPU code.
Instead of targeting general sparse computations, {\system} targets the core sparse patterns of GNNs, namely, generalized SpMM for vertex-wise computations and generalized SDDMM for edge-wise computations.
This design choice enables {\system} to fully exploit the optimization opportunities specific to GNN workloads.

\section{Conclusion}
\label{sec:conclusion}

We propose \system~to enable performant processing of graph neural network (GNN) workloads.
{\system} provides a flexible programming interface that is able to express the diverse variants of GNN workloads by composing sparse templates with customizable feature dimension computations on each vertex/edge.
{\system} extensively explores optimization opportunities in both graph traversal and feature  dimension computation.
Moreover, it decouples these two levels of optimizations to improve the productivity of developing new kernels for emerging GNN models.
\system~is portable to existing GNN frameworks as a high-performance backend.
Our evaluation verifies that \system~is comprehensive at kernel coverage and outperforms the state-of-the-art solutions.
Future work remains to utilize more intelligent tuners~\cite{autotvm} to further improve the performance, \revision{and to integrate {\system} into large-scale GNN training systems such as NeuGraph to accelerate multi-GPU training.}

\section*{Acknowledgement}
\label{sec:ack}

We thank the anonymous reviewers for valuable comments. The authors affiliated with Cornell University were funded in part by CRISP, one of six centers in JUMP, a Semiconductor Research Corporation (SRC) program sponsored by DARPA, and by AFRL and DARPA under agreement number FA8650-18-2-7863. The U.S. Government is authorized to reproduce and distribute reprints for Governmental purposes notwithstanding any copyright notation thereon. The views and conclusions contained herein are those of the authors and should not be interpreted as necessarily representing the official policies or endorsements, either expressed or implied, of AFRL and DARPA or the U.S. Government.

% \newpage

\bibliographystyle{IEEEtran}
\bibliography{main}

\begin{thebibliography}{10}
\providecommand{\url}[1]{#1}
\csname url@samestyle\endcsname
\providecommand{\newblock}{\relax}
\providecommand{\bibinfo}[2]{#2}
\providecommand{\BIBentrySTDinterwordspacing}{\spaceskip=0pt\relax}
\providecommand{\BIBentryALTinterwordstretchfactor}{4}
\providecommand{\BIBentryALTinterwordspacing}{\spaceskip=\fontdimen2\font plus
\BIBentryALTinterwordstretchfactor\fontdimen3\font minus
  \fontdimen4\font\relax}
\providecommand{\BIBforeignlanguage}[2]{{%
\expandafter\ifx\csname l@#1\endcsname\relax
\typeout{** WARNING: IEEEtran.bst: No hyphenation pattern has been}%
\typeout{** loaded for the language `#1'. Using the pattern for}%
\typeout{** the default language instead.}%
\else
\language=\csname l@#1\endcsname
\fi
#2}}
\providecommand{\BIBdecl}{\relax}
\BIBdecl

\bibitem{gnn-social-network}
Q.~Tan, N.~Liu, and X.~Hu, ``Deep representation learning for social network
  analysis,'' \emph{arXiv preprint arXiv:1904.08547}, 2019.

\bibitem{PinSage}
R.~Ying, R.~He, K.~Chen, P.~Eksombatchai, W.~L. Hamilton, and J.~Leskovec,
  ``Graph convolutional neural networks for web-scale recommender systems,''
  \emph{Int'l Conf. on Knowledge Discovery and Data Mining (KDD)}, 2018.

\bibitem{gnn-chemistry}
J.~Gilmer, S.~S. Schoenholz, P.~F. Riley, O.~Vinyals, and G.~E. Dahl, ``Neural
  message passing for quantum chemistry,'' \emph{Int'l Conf. on Machine
  Learning (ICML)}, 2017.

\bibitem{gnn-combinatorial}
Z.~Li, Q.~Chen, and V.~Koltun, ``Combinatorial optimization with graph
  convolutional networks and guided tree search,'' \emph{Conf. on Neural
  Information Processing Systems (NIPS)}, 2018.

\bibitem{neugraph}
L.~Ma, Z.~Yang, Y.~Miao, J.~Xue, M.~Wu, L.~Zhou, and Y.~Dai, ``Neugraph:
  Parallel deep neural network computation on large graphs,'' \emph{USENIX
  Annual Technical Conf. (ATC)}, 2019.

\bibitem{tensorflow}
M.~Abadi, P.~Barham, J.~Chen, Z.~Chen, A.~Davis, J.~Dean, M.~Devin,
  S.~Ghemawat, G.~Irving, M.~Isard \emph{et~al.}, ``Tensorflow: A system for
  large-scale machine learning,'' \emph{USENIX Symp. on Operating Systems
  Design and Implementation (OSDI)}, 2016.

\bibitem{PyG}
M.~Fey and J.~E. Lenssen, ``Fast graph representation learning with pytorch
  geometric,'' \emph{arXiv preprint arXiv:1903.02428}, 2019.

\bibitem{pytorch}
A.~Paszke, S.~Gross, F.~Massa, A.~Lerer, J.~Bradbury, G.~Chanan, T.~Killeen,
  Z.~Lin, N.~Gimelshein, L.~Antiga \emph{et~al.}, ``Pytorch: An imperative
  style, high-performance deep learning library,'' \emph{Conf. on Neural
  Information Processing Systems (NeurIPS)}, 2019.

\bibitem{dgl}
M.~Wang, L.~Yu, D.~Zheng, Q.~Gan, Y.~Gai, Z.~Ye, M.~Li, J.~Zhou, Q.~Huang,
  C.~Ma \emph{et~al.}, ``Deep graph library: Towards efficient and scalable
  deep learning on graphs,'' \emph{arXiv preprint arXiv:1909.01315}, 2019.

\bibitem{mkl}
Intel, ``Intel math kernel library,''
  \url{https://software.intel.com/content/www/us/en/develop/tools/math-kernel-library.html}.

\bibitem{cusparse}
Nvidia, ``Cusparse library,'' \url{https://developer.nvidia.com/cusparse}.

\bibitem{pregel}
G.~Malewicz, M.~H. Austern, A.~J. Bik, J.~C. Dehnert, I.~Horn, N.~Leiser, and
  G.~Czajkowski, ``Pregel: A system for large-scale graph processing,''
  \emph{Int'l Conf. on Management of Data (SIGMOD)}, 2010.

\bibitem{x-stream}
A.~Roy, I.~Mihailovic, and W.~Zwaenepoel, ``X-stream: Edge-centric graph
  processing using streaming partitions,'' \emph{ACM Symp. on Operating Systems
  Principles (SOSP)}, 2013.

\bibitem{powergraph}
J.~E. Gonzalez, Y.~Low, H.~Gu, D.~Bickson, and C.~Guestrin, ``Powergraph:
  Distributed graph-parallel computation on natural graphs,'' \emph{USENIX
  Symp. on Operating Systems Design and Implementation (OSDI)}, 2012.

\bibitem{ligra}
J.~Shun and G.~E. Blelloch, ``Ligra: A lightweight graph processing framework
  for shared memory,'' \emph{ACM SIGPLAN Notices}, 2013.

\bibitem{gunrock}
Y.~Wang, A.~Davidson, Y.~Pan, Y.~Wu, A.~Riffel, and J.~D. Owens, ``Gunrock: A
  high-performance graph processing library on the gpu,'' \emph{ACM SIGPLAN
  Notices}, 2016.

\bibitem{flashgraph}
D.~Zheng, D.~Mhembere, R.~Burns, J.~Vogelstein, C.~E. Priebe, and A.~S. Szalay,
  ``Flashgraph: Processing billion-node graphs on an array of commodity ssds,''
  \emph{USENIX Conf. on File and Storage Technologies (FAST)}, 2015.

\bibitem{cagra}
Y.~Zhang, V.~Kiriansky, C.~Mendis, S.~Amarasinghe, and M.~Zaharia, ``Making
  caches work for graph analytics,'' \emph{IEEE Int'l Conf. on Big Data}, 2017.

\bibitem{gridgraph}
X.~Zhu, W.~Han, and W.~Chen, ``Gridgraph: Large-scale graph processing on a
  single machine using 2-level hierarchical partitioning,'' \emph{USENIX Annual
  Technical Conf. (ATC)}, 2015.

\bibitem{simd-efficient-graph-gpu}
F.~Khorasani, R.~Gupta, and L.~N. Bhuyan, ``Scalable simd-efficient graph
  processing on gpus,'' \emph{Int'l Conf. on Parallel Architectures and
  Compilation Techniques (PACT)}, 2015.

\bibitem{cusha}
F.~Khorasani, K.~Vora, R.~Gupta, and L.~N. Bhuyan, ``Cusha: Vertex-centric
  graph processing on gpus,'' \emph{Int'l Symp. on High-Performance Parallel
  and Distributed Computing (HPDC)}, 2014.

\bibitem{santoro2017simple}
A.~Santoro, D.~Raposo, D.~G. Barrett, M.~Malinowski, R.~Pascanu, P.~Battaglia,
  and T.~Lillicrap, ``A simple neural network module for relational
  reasoning,'' \emph{Conf. on Neural Information Processing Systems (NIPS)},
  2017.

\bibitem{palm2018recurrent}
R.~Palm, U.~Paquet, and O.~Winther, ``Recurrent relational networks,''
  \emph{Conf. on Neural Information Processing Systems (NIPS)}, 2018.

\bibitem{chen2018tvm}
T.~Chen, T.~Moreau, Z.~Jiang, L.~Zheng, E.~Yan, H.~Shen, M.~Cowan, L.~Wang,
  Y.~Hu, L.~Ceze \emph{et~al.}, ``{TVM}: An automated end-to-end optimizing
  compiler for deep learning,'' \emph{USENIX Symp. on Operating Systems Design
  and Implementation (OSDI)}, 2018.

\bibitem{graphnets}
P.~W. Battaglia, J.~B. Hamrick, V.~Bapst, A.~Sanchez-Gonzalez, V.~Zambaldi,
  M.~Malinowski, A.~Tacchetti, D.~Raposo, A.~Santoro, R.~Faulkner
  \emph{et~al.}, ``Relational inductive biases, deep learning, and graph
  networks,'' \emph{arXiv preprint arXiv:1806.01261}, 2018.

\bibitem{gcn}
T.~N. Kipf and M.~Welling, ``Semi-supervised classification with graph
  convolutional networks,'' \emph{arXiv preprint arXiv:1609.02907}, 2016.

\bibitem{gat}
P.~Veli{\v{c}}kovi{\'c}, G.~Cucurull, A.~Casanova, A.~Romero, P.~Lio, and
  Y.~Bengio, ``Graph attention networks,'' \emph{arXiv preprint
  arXiv:1710.10903}, 2017.

\bibitem{agnn}
K.~K. Thekumparampil, C.~Wang, S.~Oh, and L.-J. Li, ``Attention-based graph
  neural network for semi-supervised learning,'' \emph{arXiv preprint
  arXiv:1803.03735}, 2018.

\bibitem{vaswani2017attention}
A.~Vaswani, N.~Shazeer, N.~Parmar, J.~Uszkoreit, L.~Jones, A.~N. Gomez,
  {\L}.~Kaiser, and I.~Polosukhin, ``Attention is all you need,'' \emph{Conf.
  on Neural Information Processing Systems (NIPS)}, 2017.

\bibitem{sddmm-2}
H.~Zhao, ``High performance machine learning through codesign and rooflining,''
  Ph.D. dissertation, UC Berkeley, 2014.

\bibitem{ragan2013halide}
J.~Ragan-Kelley, C.~Barnes, A.~Adams, S.~Paris, F.~Durand, and S.~Amarasinghe,
  ``Halide: A language and compiler for optimizing parallelism, locality, and
  recomputation in image processing pipelines,'' \emph{ACM SIGPLAN Notices},
  2013.

\bibitem{hilbert}
F.~McSherry, M.~Isard, and D.~G. Murray, ``Scalability! but at what cost?''
  \emph{Workshop on Hot Topics in Operating Systems (HotOS)}, 2015.

\bibitem{yang2018design}
C.~Yang, A.~Bulu{\c{c}}, and J.~D. Owens, ``Design principles for sparse matrix
  multiplication on the gpu,'' \emph{European Conf. on Parallel Processing
  (Euro-Par)}, 2018.

\bibitem{tree-reduce}
M.~Harris \emph{et~al.}, ``Optimizing parallel reduction in cuda,''
  \url{http://developer.download.nvidia.com/assets/cuda/files/reduction.pdf},
  2012.

\bibitem{aws-cnn-cpu}
Y.~Liu, Y.~Wang, R.~Yu, M.~Li, V.~Sharma, and Y.~Wang, ``Optimizing cnn model
  inference on cpus,'' \emph{USENIX Annual Technical Conf. (ATC)}, 2019.

\bibitem{aws-cnn-gpu}
L.~Wang, Z.~Chen, Y.~Liu, Y.~Wang, L.~Zheng, M.~Li, and Y.~Wang, ``A unified
  optimization approach for cnn model inference on integrated gpus,''
  \emph{Int'l Conf. on Parallel Processing (ICPP)}, 2019.

\bibitem{opentuner}
J.~Ansel, S.~Kamil, K.~Veeramachaneni, J.~Ragan-Kelley, J.~Bosboom, U.-M.
  O'Reilly, and S.~Amarasinghe, ``Opentuner: An extensible framework for
  program autotuning,'' \emph{Int'l Conf. on Parallel Architectures and
  Compilation Techniques (PACT)}, 2014.

\bibitem{autotvm}
T.~Chen, L.~Zheng, E.~Yan, Z.~Jiang, T.~Moreau, L.~Ceze, C.~Guestrin, and
  A.~Krishnamurthy, ``Learning to optimize tensor programs,'' \emph{Conf. on
  Neural Information Processing Systems (NIPS)}, 2018.

\bibitem{minigun}
``Minigun: Light-weight gpu kernel interface for graph operations,''
  \url{https://github.com/dglai/minigun}, 2019.

\bibitem{graphsage}
W.~Hamilton, Z.~Ying, and J.~Leskovec, ``Inductive representation learning on
  large graphs,'' \emph{Conf. on Neural Information Processing Systems (NIPS)},
  2017.

\bibitem{graphblas}
J.~Kepner, P.~Aaltonen, D.~Bader, A.~Bulu{\c{c}}, F.~Franchetti, J.~Gilbert,
  D.~Hutchison, M.~Kumar, A.~Lumsdaine, H.~Meyerhenke \emph{et~al.},
  ``Mathematical foundations of the graphblas,'' \emph{IEEE High Performance
  Extreme Computing Conf. (HPEC)}, 2016.

\bibitem{graphmat}
N.~Sundaram, N.~Satish, M.~M.~A. Patwary, S.~R. Dulloor, M.~J. Anderson, S.~G.
  Vadlamudi, D.~Das, and P.~Dubey, ``Graphmat: High performance graph analytics
  made productive,'' \emph{Int'l Conf. on Very Large Data Bases (VLDB)}, 2015.

\bibitem{graphlinearalgebra}
J.~Kepner and J.~Gilbert, \emph{Graph algorithms in the language of linear
  algebra}.\hskip 1em plus 0.5em minus 0.4em\relax Society for Industrial and
  Applied Mathematics, 2011.

\bibitem{semispmm}
D.~Zheng, D.~Mhembere, V.~Lyzinski, J.~T. Vogelstein, C.~E. Priebe, and
  R.~Burns, ``Semi-external memory sparse matrix multiplication for
  billion-node graphs,'' \emph{IEEE Trans. on Parallel and Distributed Systems
  (TPDS)}, 2016.

\bibitem{taco}
F.~Kjolstad, S.~Kamil, S.~Chou, D.~Lugato, and S.~Amarasinghe, ``The tensor
  algebra compiler,'' \emph{Object-Oriented Programming, Systems, Languages,
  and Applications (OOPSLA)}, 2017.

\end{thebibliography}

\end{document}